\documentclass{article}

\usepackage{arxiv}

\usepackage[utf8]{inputenc} 
\usepackage[T1]{fontenc}    
\usepackage{hyperref}       
\usepackage{url}            
\usepackage{booktabs}       
\usepackage{amsfonts}       
\usepackage{nicefrac}       
\usepackage{microtype}      
\usepackage{lipsum}		
\usepackage{graphicx}
\usepackage[numbers]{natbib}
\usepackage{doi}

\usepackage[dvipsnames]{xcolor}

\usepackage{amsmath}
\usepackage{graphicx}
\usepackage{booktabs}
\usepackage{multirow}

\usepackage{pifont}
\newcommand{\cmark}{\ding{51}}
\newcommand{\xmark}{\ding{55}}

\title{Benchmarking M-LTSF: Frequency and Noise-Based Evaluation of Multivariate Long Time Series Forecasting Models}

\date{} 					

\author{%
  Nick Janssen\\
  Institute of Information Processing\\
  Leibniz University Hanover\\
  \texttt{janssen@tnt.uni-hannover.de} \\
  \And
  Melanie Schaller\\
  Institute of Information Processing\\
  Leibniz University Hanover\\
  \texttt{schaller@tnt.uni-hannover.de} \\
  \And
  Bodo Rosenhahn\\
  Institute of Information Processing\\
  Leibniz University Hanover\\
  \texttt{rosenhahn@tnt.uni-hannover.de} \\
}

\renewcommand{\shorttitle}{Benchmarking M-LTSF}

\hypersetup{
pdftitle={Benchmarking M-LTSF: Frequency and Noise-Based Evaluation of Multivariate Long Time Series Forecasting Models},
pdfauthor={Nick Janssen, Melanie Schaller, Bodo Rosenhahn},
pdfkeywords={Multivariate Long Time Series Forecasting, Time series analysis, Signal Processing, Noise Analysis, Spectral Analysis},
}

\begin{document}
\maketitle

\begin{abstract}
Understanding the robustness of deep learning models for multivariate long-term time series forecasting (M-LTSF) remains challenging, as evaluations typically rely on real-world datasets with unknown noise properties.
We propose a simulation-based evaluation framework that generates configurable synthetic dataset instances based on controllable signal components, noise types, signal-to-noise ratios, and frequency characteristics.
These configurable components aim to model real-world multivariate time series data without the ambiguity of unknown noise.
This framework enables fine-grained, systematic evaluation of M-LTSF models under controlled and diverse scenarios.
We benchmark four representative architectures S-Mamba, iTransformer, R-Linear, and Autoformer.
Our analysis shows that model performance deteriorates sharply when lookback windows fail to capture full seasonal cycles. 
S-Mamba and Autoformer perform best on sawtooth patterns, whereas R-Linear and iTransformer prefer sinusoidal signals. 
White and Brownian noise consistently impair performance at lower signal-to-noise ratios.
Additionally, S-Mamba is sensitive to trend noise, and iTransformer to seasonal noise. 
Spectral analysis further indicates that S-Mamba and iTransformer offer superior frequency reconstruction.
This controlled approach, based on our synthetic evaluation framework, offers deeper insights into model-specific strengths and limitations through the aggregation of MSE scores and provides concrete guidance for model selection based on signal characteristics and noise conditions.
\end{abstract}

\keywords{Multivariate Long Time Series Forecasting (M-LTSF) \and Time series analysis \and Signal Processing \and Noise Analysis \and Spectral Analysis}

\section{Introduction}
Time series forecasting plays a crucial role across diverse fields such as energy systems~\cite{zhou2022_fedformer,forootani2024_climateawaredeepneural,mustafa2025_energyforecasting}, meteorology~\cite{sun2023_stecformer,bouallegue2024_weatherforecast}, traffic flow modeling~\cite{liu2023_largest,liu2025_traffic} or the modeling of sensor networks~\cite{schaller2025_s4dbio,farahani2025_manufacturingsystems}. 
Reliable forecasts support proactive decision-making, effective risk management, and efficient planning.
As high-resolution temporal data becomes increasingly available, the need for robust and scalable forecasting models has grown more important than ever.\\
A time series represents data points ordered in time and can be categorized as either univariate, when consisting of a single variable, or multivariate, when involving multiple interdependent variables~\cite{aboagyesarfo2015_unimultitsf}. 
Correspondingly, forecasting techniques are typically divided into Univariate Time Series Forecasting (U-TSF) and Multivariate Time Series Forecasting (M-TSF).
Another important distinction lies in the forecasting horizon: traditional (short-horizon) forecasting aims to predict only the immediate future steps, whereas Long-Term Time Series Forecasting (LTSF) focuses on generating accurate predictions over extended horizons, which is often more challenging due to error accumulation and complex temporal dependencies~\cite{chen2023_ltsfsurvey}.\\
Classical statistical models such as Autoregressive Integrated Moving Average (ARIMA)~\cite{fattah2018_arima,ewa2021_arima} for univariate data and Vector Autoregression (VAR)~\cite{triantafyllopoulos2008_var} for multivariate data have long been regarded as foundational methods. 
Subsequently, machine learning approaches, particularly decision tree–based methods~\cite{chen2016_xgboost,zhai2020_xgboosttsf}, demonstrated enhanced predictive performance compared to traditional statistical models. 
More recently, deep learning architectures ranging from Transformer-based methods~\cite{wu2021_autoformer,zhou2021_informer,liu2023_itransformer}, state-space models~\cite{gu2021_s4,gu2023_mamba,wang2025_smamba} or simple linear approaches~\cite{zeng2023_dlinear,li2023_rlinear} to larger foundational models~\cite{ansari2024_chronos,woo2024_moirai,das2024_timesfm}, have achieved state-of-the-art results across a wide range of datasets.\\
\begin{figure}[h]
\begin{center}
\includegraphics[width=0.8\columnwidth]{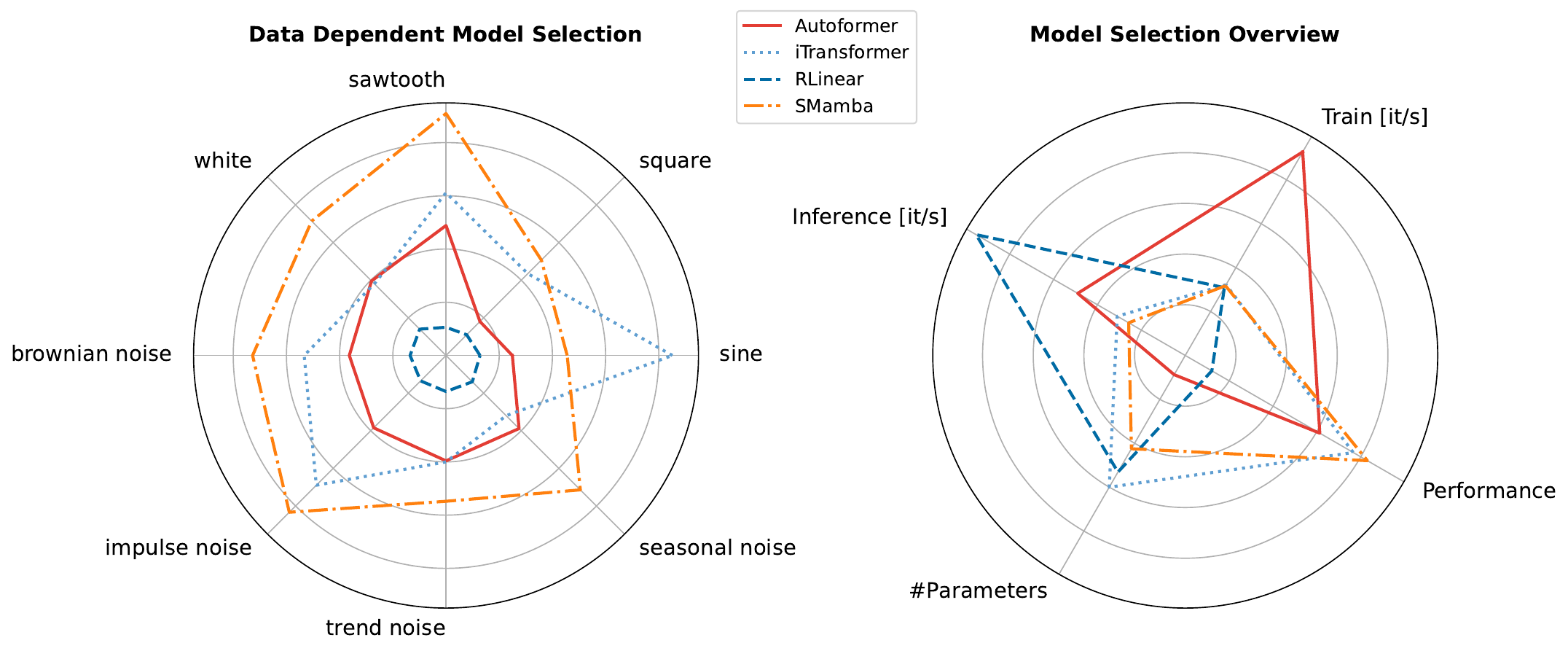}
\caption{Radarplots to guide model selection. Left Figure inverse of the best MSE value of each model across seasonality types (sine, square, sawtooth) and noise types (white noise, Brownian noise, impulse noise, trend noise and seasonal noise) with an SNR value of 100. S-Mamba shows best overall performance on all different dataset characteristics. The right figure shows the efficiency of the model. Features are inference (Iterations/s), training time (It/s), number of parameters (inverse) and the best MSE Score (inverse) across all evaluated experiments. R-Linear shows fast inference time but lacks in performance, while iTransformer and S-Mamba show good performance but worse inference speed.}
\label{fig:model_selection}
\end{center}
\end{figure}%
Equally important is the in-depth evaluation of these models, which is essential for understanding their decision-making processes, ensuring transparency, and improving trust in their predictions.
Current research mainly focuses on achieving superior performance on established benchmark datasets, with success typically measured through metrics such as Mean Squared Error (MSE) or Mean Absolute Error (MAE)~\cite{wu2021_autoformer,zhou2021_informer,zhou2022_fedformer}.
While these measures of performance yield algorithmic innovation, it provides limited understanding of the fundamental capabilities and limitations of these models.
When relying on metrics such as MSE without accounting for dataset characteristics, the evaluation predominantly measures the direct fit to the benchmark time series, with the risk of not distinguishing between learning the true underlying signal and overfitting to noise.\\
A big challenge in current M-LTSF evaluation lies in the ambiguous nature of real-world benchmark datasets~\cite{liu2023_largest, ran2025_hrextremeweather}. 
These datasets inherently contain unknown combinations of signal components, noise characteristics, and other temporal patterns, making it impossible to isolate and analyze specific model behaviors without prior knowledge of the underlying system, data acquisition and preparation. 
Recent benchmarking efforts such as TFB~\cite{qiu2024_tfb} and GIFT-Eval~\cite{gervet2024_gifteval} have addressed the need for fair and comprehensive evaluation across diverse datasets and settings, though without explicit characterization of noise properties.
Consequently, critical questions remain, e.g. model performance across different frequency ranges, their robustness to various noise types, the extent to which performance degrades with changing signal-to-noise ratio (SNR), and the impact of input window size on model capabilities.\\
Our work addresses these fundamental questions by introducing a controlled evaluation framework based on a parameterizable synthetic multivariate time series dataset. 
Our approach enables systematic analysis of model behavior under controlled conditions, providing insights that are difficult or impossible to obtain through traditional benchmark evaluation. 
As illustrated in Figure \ref{fig:model_selection}, these insights include informed model selection for specific datasets when signal and noise characteristics are known, as well as potential architectural improvements for M-LTSF models based on frequency-specific performance patterns and noise robustness findings.
We evaluate four representative M-LTSF models, including S-Mamba~\cite{wang2025_smamba} as a state-space model approach, iTransformer~\cite{liu2023_itransformer} representing attention-based designs, R-Linear~\cite{li2023_rlinear} as a linear method, and Autoformer~\cite{wu2021_autoformer} utilizing a decomposition-based design, across multiple combinations of signal and noise variations, revealing previously unexplored aspects of model performance and robustness.\\
\textbf{Our key contributions are:}
\begin{itemize}
    \item \textbf{Parameterizable synthetic dataset framework:} We develop a hierarchical generation approach to simulate real-world multivariate time series with precise control over signal characteristics and noise types, providing a reproducible benchmark for systematic evaluation of model performance.
    \item \textbf{Model Selection Guidance:} We showcase how our framework can guide model selection based on dataset-specific characteristics. In our evaluation, S-Mamba achieves the highest overall accuracy but suffers from slower inference times, whereas R-Linear offers fast inference but its performance declines on datasets with a large number of variates.
    \item \textbf{Characterization of model behavior under varying signals and noise:} We provide a framework to analyze how different architectures respond to diverse signal patterns and noise types. This allows systematic exploration of model preferences, highlighting, for example, which architectures are more suited to certain seasonalities or exhibit sensitivity to specific noise conditions, thereby offering insights into their strengths and vulnerabilities without focusing on specific performance outcomes.
    \item \textbf{Codebase Publication:} The complete framework for the synthetic dataset benchmark are accessible open-source on GitHub.\footnote{Code will be released after acceptance.}
\end{itemize}

\section{Related Work}
\noindent Multivariate Long Time Series Forecasting (M-LTSF) represents a fundamental challenge in temporal data analysis, where the objective is to predict future values of multiple interconnected time series variables based on their historical observations.
Formally, given a multivariate time series $\mathbf{X_{1:T}} = \{\mathbf{x_1}, \mathbf{x_2}, \mathbf{x_3}, ..., \mathbf{x_T}\}$ with lookback horizon $T$, where $\mathbf{x_k} \in \mathbb{R}^V$ represents $V$ variates at time step $k$, the goal is to predict future values $\mathbf{X_{T+1:T+H}} = \{\mathbf{x_{T+1}}, \mathbf{x_{T+2}}, ..., \mathbf{x_{T+H}}\}$ over a forecast horizon $H$~\cite{hsu1998_tslcrisis,zhou2021_informer}.
Multivariate Long Time-Series Forecasting (M-LTSF) is defined based on the relationship between prediction horizons and data cycles. 
While traditional definitions characterize M-LTSF as forecasting “a more distant future”, a more precise definition emerges from industry practices across various domains~\cite{manibardo2022_roadtrafficforecasting}. 
The definition of M-LTSF is typically considered when the prediction horizon approaches or exceeds the maximum cycle present in the data. 
For instance, in finance, forecasting over 10 years with economic cycles of 2 to 10 years~\cite{harvey2007_economictimeseries} aligns with the definition of M-LTSF, while in traffic predictions, several days may be sufficient~\cite{hao2018_forcastingtrafficflow}. 
This cycle-based criterion provides a domain-specific representative definition for M-LTSF~\cite{chen2023_ltsfsurvey}.

\subsection{Long Time Series Forecasting Architectures}
The multivariate aspect introduces additional complexity by requiring models to learn both temporal patterns within individual series and cross-series dependencies that may involve varying time delays and nonlinear dynamics that change over time.
This multidimensional challenge, combined with the extended temporal scope of M-LTSF, creates significant computational and modeling difficulties that demand specialized architectural approaches.\\
\textbf{Transformer-based Architectures:} 
The success of Transformers~\cite{vaswani2023_transformer} architectures in natural language processing and computer vision led to their adoption for time series forecasting~\cite{zhou2021_informer}, where models leverage attention mechanisms to capture long-range dependencies or cross-series correlations. 
This has given rise to several specialized variants tailored to the unique challenges of time series data.
\textit{Autoformer}~\cite{wu2021_autoformer} introduced a decomposition-based approach that explicitly separates trend and seasonal components by computing a running mean across the time dimension, helping the model better understand the structures of the time series. It also employs an autocorrelation that outperforms self-attention, improving the accuracy of forecasts.
\textit{iTransformer}~\cite{liu2023_itransformer} in contrast takes a simpler yet effective approach by rethinking the input representation. 
Instead of applying attention across time steps, it applies self-attention across variates, treating each features time series as a token. 
This design improves multivariate correlation modeling, handles longer input sequences more efficiently. 
Despite these innovations, Transformer-based models face fundamental challenges in time series forecasting. 
The self-attention mechanism is permutation-invariant. which can obscure the temporal order that is crucial for forecasting tasks~\cite{lim2021_film, wu2021_autoformer}.
Additionally, unlike language data, numerical time series lack rich semantic relationships, limiting the effectiveness of attention mechanisms~\cite{zeng2023_dlinear, liu2023_itransformer}. \\
\textbf{Linear Models:} 
These limitations of Transformer-based approaches led to a surprising research shift with the introduction of simple linear models for M-LTSF using types of instance normalization~\cite{kim2022_revin}. 
An in-depth analysis by Zeng et~al.~\cite{zeng2023_dlinear} revealed that most Transformer architectures fail to extract temporal relations from long sequences, with forecasting errors often not reducing or increasing with larger look-back windows or when predicting on the previous lookback window.
These findings showed that the attention mechanism only captures similar temporal information from the adjacent time series sequence but not the underlying trend and periodicity which questioned the assumption that the more complex models are better suited for temporal data~\cite{liu2023_nonstationarytransformers}.
However, linear models also face specific challenges in multivariate settings, particularly when dealing with multiple periods among channels. 
Li et~al.~\cite{li2023_rlinear} introduced the \textit{R-Linear} architecture and showed that, while linear mapping can effectively capture periodicity in time series, it struggles with multi-channel datasets characterized by heterogeneous periodic patterns.
Channel Independent (CI) modeling proposed by R-Linear addresses this limitation by treating each channel separately. 
While CI modeling improves forecasting accuracy, it significantly increases computational overhead, scaling linearly with the number of channels.\\
\textbf{State Space Models:} 
Another emerging approach to time series forecasting involves modeling the underlying system using a state space model. 
The core model proposed by Gu et~al.~\cite{gu2021_s4} (\textit{S4}) efficiently handles variable-length sequences while maintaining linear complexity with respect to sequence length thus giving them an advantage over Transformers. 
Expanding on this foundation, the \textbf{Mamba} architecture of Gu et~al.~\cite{gu2023_mamba} introduces a novel input-dependent, time-varying state space model that enables efficient processing of long sequences.
Unlike traditional models, Mamba dynamically adapts its state transitions based on the input, allowing it to capture complex temporal dependencies. 
Its built-in selective mechanism may facilitate attention to relevant historical information while filtering out irrelevant or noisy data.
Building upon this foundation, the \textit{S-Mamba} architecture~\cite{wang2025_smamba} adapts Mamba's selective state space mechanism specifically for time series forecasting by replacing the attention computation in the iTransformer architecture with bidirectional Mamba blocks. This substitution maintains the pattern recognition capabilities while achieving near-linear computational complexity, making it particularly suitable for efficient multivariate time series prediction.

\subsection{Benchmark Datasets}
Recent research on M-LTSF methods is commonly based on a set of standard benchmark datasets for evaluation.
The ETT (Electricity Transformer Temperature) dataset presented by Wu et~al.~\cite{wu2021_autoformer} contains 2-year electricity transformer data from China with variants for hourly and 15-minute sampling rates. 
Other commonly used datasets include electricity consumption data from 321 consumers, daily exchange rates from eight countries, hourly traffic occupancy rates from California freeways, and meteorological data with 21 indicators recorded every 10 minutes ~\cite{zhou2021_informer,lai2018_longshorttermtemporalpatterns}.\\
While these benchmarks facilitate model comparison across datasets with varying numbers of variables and different temporal frequencies, they present significant limitations for noise analysis. 
The datasets lack detailed documentation of noise characteristics, measurement errors, sensor reliability, or data preprocessing steps. 
This absence of noise characterization makes it difficult to assess whether performance differences stem from algorithmic improvements or varying noise sensitivities, thereby limiting their utility for rigorous robustness evaluation.
Moreover, without precise information about the noise, it remains unclear whether certain effects cancel each other out. 
A model that performs well under one noise condition might underperform under another, potentially leading to uncertain or misleading conclusions about the relative superiority of models.

\subsection{Noise analysis and modeling in M-LTSF}
A critical challenge in M-LTSF is the tendency of deep learning methods to overfit on noise present in the training data, particularly when the signal-to-noise ratio is low~\cite{lim2021_film}.
Traditional approaches to noise analysis in time series forecasting rely on oversimplified assumptions. 
Singh~\cite{singh1999_noise} evaluated noise impact by only injecting Gaussian noise and filtering based on a cutoff frequency that classified low-frequency content as "trend" energy and high frequencies as noise. 
This approach is problematic because real-world noise is rarely only Gaussian in nature, and the binary classification of frequency components oversimplifies the complex spectral characteristics of time series, where meaningful patterns can exist across the entire frequency spectrum.
Further, recent works on probabilistic forecasting~\cite{yoon2022_probforecasting} and noise robustness~\cite{dey2025_hadl,ewa2021_arima} have shown that models vary in their ability to maintain performance under noisy or adversarial conditions.
Wang et~al.~\cite{wang2025_syntheticts} demonstrated the value of synthetic data for controlled evaluation of Transformer architectures across ten signal types, though their analysis was limited to Transformer-based models and combined multiple noise types simultaneously rather than isolating individual noise characteristics.

\section{M-LTSF Benchmark Framework}
To investigate the behavior of M-LTSF models under controlled and interpretable conditions, we developed a parameterizable synthetic dataset framework specifically designed for multivariate long time series forecasting. 
This framework provides precise control over both signal and noise characteristics, enabling systematic evaluation of model performance across diverse scenarios.
The dataset construction follows a hierarchical component-based approach, as illustrated in Figure \ref{fig:component_aggregation}. 
Each time series is composed of multiple predefined signal and noise components that are systematically distributed across variates to create realistic cross-series dependencies. 
The exact number of instances for each component type (e.g., trends, seasonalities, cycles, or specific noise patterns) can be specified in advance, providing fine-grained control over the dataset composition.
The framework employs a probabilistic assignment mechanism that balances diversity and uniformity: components can be shared across multiple variates to model inter-dependencies, while a penalty-based selection process prevents excessive concentration of components within individual variates.
Specifically, when assigning components to variates, the selection probability $p(v)$ for variate $v \in [1, V]$ is:
\begin{equation}
    p(v) = \frac{s_{v}^{-p}}{\sum_{j=1}^{V} s_{j}^{-p}}
\end{equation}
where $s_v$ represents the number of components already assigned to variate $v$, and the configurable exponent $p$ controls the strength of the penalty for previous selections. 
This mechanism ensures that the resulting dataset maintains both meaningful cross-variate relationships and balanced component distribution.\\
The construction process ultimately produces, for each variate, a dedicated list of signal components and a list of noise components, from which the final time series are synthesized through component aggregation.

\subsection{Signal Components}
\begin{figure}[htp]
\begin{center}
\includegraphics[width=0.7\columnwidth]{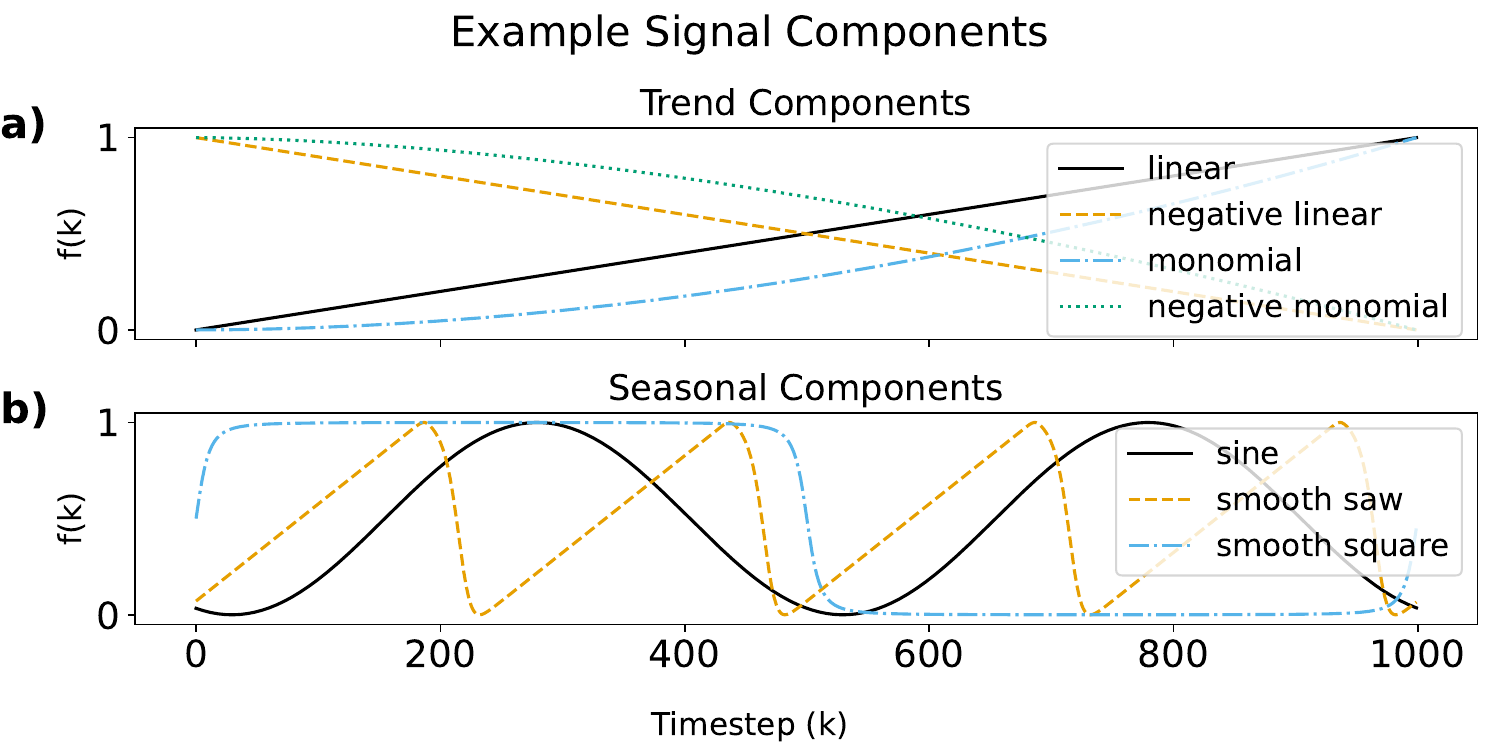}
\caption{Example synthetic time series components. Upper panel: trend components with varying exponent values $b$. Lower panel: seasonality components showing different waveform types (Sine, Saw, Square) with varying frequency and phase parameters.}
\label{fig:signal_components}
\end{center}
\end{figure}

The signal components as shown in Figure \ref{fig:signal_components} are the building blocks to create the time series for the underlying system and are split into trend and seasonality components. All following types have the function definition of type $g : \mathbb{Z} \to \mathbb{R}$ since the dataset is discrete in the time domain.\\
\textbf{Trend Component:} 
Trend patterns arise from factors such as sensor drift or long-term changes like the global average temperature which are modeled using monomial functions:
\begin{equation}
    \quad g(k) = a \cdot k^b
\end{equation}
where the parameter $a \in [-1, 1]$ accounts for the inverse trend functionality. 
The power parameter $p$ is sampled from a uniform distribution with specifiable range. 
This formulation allows for both linear ($b=1$) and non-linear trend patterns.\\
\textbf{Seasonal Component:} 
For seasonal patterns of the signal three different types of periodic patterns can be randomly sampled. For the following type definitions applies, $x_k = 2 \pi f \cdot k + \phi$, where $f$ is the frequency and $\phi$ is the phase. 
\begin{equation}
    \phi \sim \mathcal{U}(0, 2\pi), \quad f \sim \mathcal{N}(\mu, \sigma^2) \text{ limited to } [f_{\min}, f_{\max}]
\end{equation}
The phase $\phi$ is sampled from a uniform distribution between $[0, 2\pi]$ and the frequency is sample from a bounded Gaussian distribution.
\begin{itemize}
    \item 1. Sinusoidal: Represents smooth oscillations, for example daily temperature cycles or circadian rhythms.
    \begin{equation}
        g(k) = \sin(x_k)
    \end{equation}
    \item 2. Smooth Sawtooth: Models gradual increases followed by sharp drops, as observed in charging–discharging cycles of batteries.
    \begin{equation}
        g(k) = \arcsin(\tanh(10 * \cos(x_k)) * \sin(x_k))
    \end{equation}
    \item 3. Smooth Square: Captures flattened or saturated oscillations, similar to signals constrained by physical limits, such as amplitude-limited electronic signals or flow controls in industrial fluid systems.
    \begin{equation}
        g(k) = \frac{\sin(x_k)}{\sqrt{0.005 + \sin(x_k)^2}}
    \end{equation}
\end{itemize}
These types of seasonal patterns create the possibility for complex spectral characteristics with harmonic structures.

\subsection{Noise Components}
\begin{figure}[htp]
\begin{center}
\includegraphics[width=0.7\columnwidth]{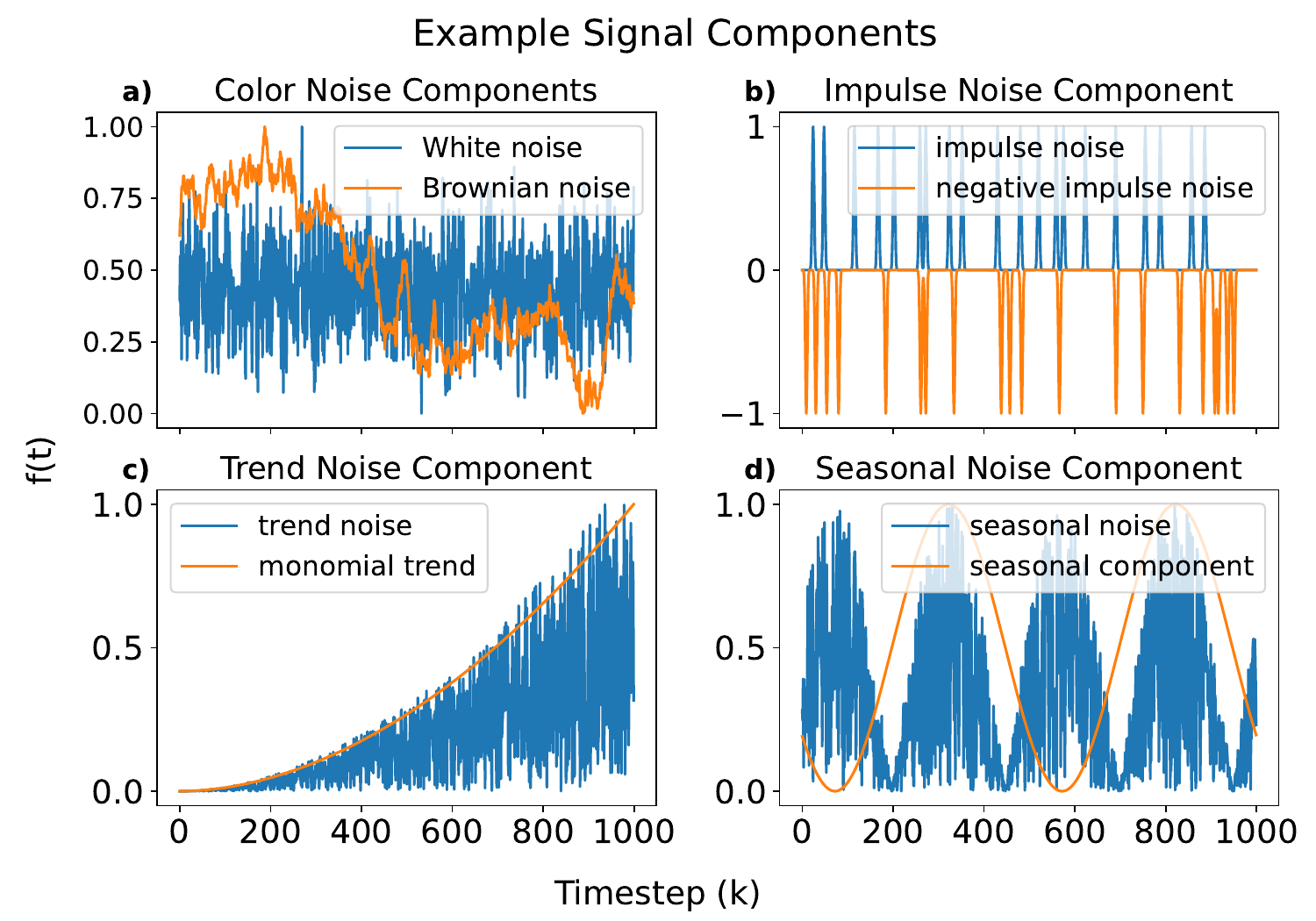}
\caption{Example time series demonstrating different noise types: White Noise and Brownian Noise as the cumulative of White Noise (top left), Impulse Noise (top right), Trend Noise (bottom left), and Seasonal Noise (bottom right). All noise types except white and Brownian noisecan exhibit inverted or anti-proportional characteristics.}
\label{fig:noise_components}
\end{center}
\end{figure}%
The following noise components~\cite{boyat2015_noisemodels} are categorized into two groups, signal independent and signal dependent enabling the creation of a complex noise characteristic (Figure \ref{fig:noise_components}).

\subsubsection{Signal Independent Noise}
The independent noise components introduce disturbances that are unrelated to the underlying signal. 
These simulate scenarios where external random fluctuations or sensor errors affect the data in a non-systematic manner.\\
\textbf{White Noise:} 
In a classic model for random error, white noise is sampled from a standard normal distribution. 
\begin{equation}
    W(k) \sim \mathcal{N}(0, 1)
\end{equation}
It has zero mean, unit variance, and no autocorrelation, meaning each time step is independent of the others. 
This models stochastic disturbances with no temporal structure.\\
\textbf{Brownian Noise:} 
Brownian noise, also called red noise, can be constructed as the cumulative sum of white noise. 
\begin{equation}
    W(k) = \sum_{i=1}^{k} \epsilon_i, \quad \epsilon_i \sim \mathcal{N}(0,1)
\end{equation}
It has zero mean, but its variance grows linearly with time, and successive values are highly correlated. 
This models stochastic disturbances with strong temporal structure, where each step depends on the accumulation of past random shocks.\\
\textbf{Impulse Noise:} 
Impulse noise consists of a number of $N_{imp}$ sporadic, high-amplitude spikes that occur at random time steps.
Each impulse is modeled as a Gaussian function centered at a random location $\mu_i$ with a randomly sampled width $\sigma^2$ and amplitude $a$ based on the specified configuration resulting in the impulse noise time series $W(k)$.
\begin{equation}
    W(k) = \sum_{i=1}^{N_{imp}} \frac{a}{\sqrt{2\pi\sigma^2}} \cdot \exp \Biggl(-\frac{(k - \mu_i)^2}{2\sigma^2} \Biggr)
\end{equation}
This mimics transient anomalies such as hardware glitches, data transmission errors, or external shocks.

\subsubsection{Signal Dependent Noise}
The dependent noise components are explicitly designed to correlate with the signal itself. 
They simulate noise that arises from imperfect measurement or system dynamics, where the amount of noise is modulated by the underlying signal's properties.\\
\textbf{Trend Noise:} 
This noise type scales proportionally with the magnitude of the trend signal component. 
In practice, the noise amplitude at each time step is determined by multiplying a sample from a white noise process with the corresponding value of the trend signal.
\begin{equation}
    W(k) = a \cdot x_k^{\text{trend}} \cdot \gamma_k, \quad \gamma_k \sim \mathcal{N}(0, 1), \quad a \in \{-1, 1\}
\end{equation}
This reflects situations where long-term drift in the signal increases the uncertainty of measurements or predictions.\\
\textbf{Seasonal Noise:} 
Analogous to trend noise, seasonal noise is modulated by the amplitude of the seasonal component. 
\begin{equation}
    W(k) = a \cdot x_k^{\text{seasonal}} \cdot \gamma_k, \quad \gamma_k \sim \mathcal{N}(0, 1), \quad a \in \{-1, 1\}
\end{equation}
For example, in meteorological or energy datasets, measurement noise might be more prominent during peak seasonal cycles due to high volatility. 
The noise amplitude is shaped by the underlying seasonal signal, leading to temporally structured noise that varies in intensity across each cycle.

\subsection{Multivariate Time Series Generation}
\begin{figure}[htp]
\begin{center}
\includegraphics[width=0.8\columnwidth]{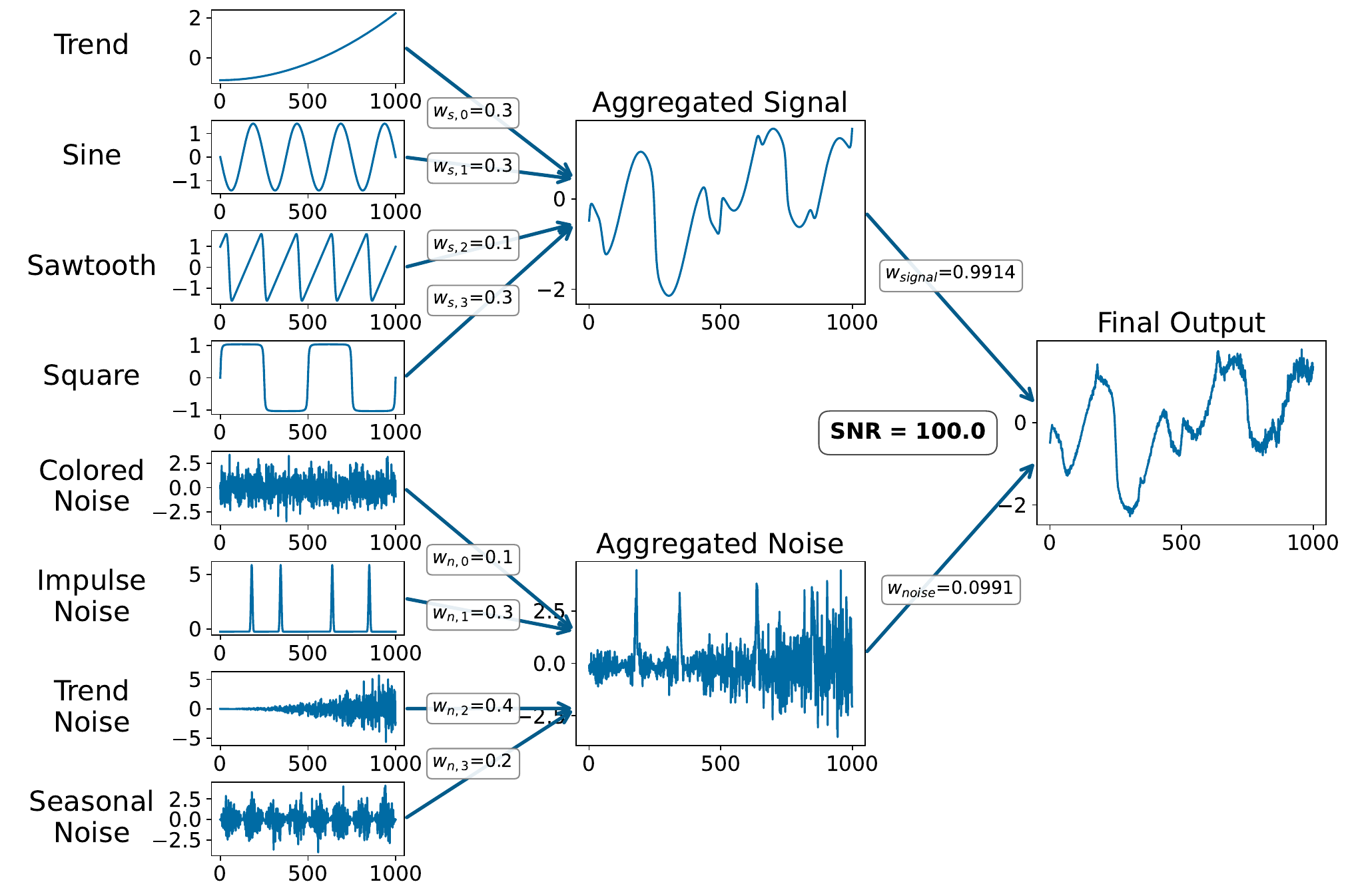}
\caption{Overview of time series generation for a univariate signal. A time series is constructed for each component individually. Each component is weighted by a randomly sampled factor ($w_{signal,i}$ and $w_{noise,i}$) and aggregated into separate signal and noise time series. The signal and noise are then scaled according to the specified signal-to-noise ratio and combined to produce the final time series shown on the right. All displayed time series are z-normalized, and the shown weights are randomly sampled examples.}
\label{fig:component_aggregation}
\end{center}
\end{figure}%
To create the time series based on the sampled components for each variate $k \in K$, the signal and noise are constructed in two stages. 
First, signal components are combined into a single signal time series, and noise components are combined into a single noise time series:\\
For both signal ($y_{\text{signal}}^{(k)}$) and noise ($y_{\text{noise}}^{(K)}$) time series the components individual time series are scaled by a random sampled weight $w_i$, with all weights summing to one.
\begin{align}
    \tilde{w}_{\text{signal},i}^{(k)} &\sim \mathcal{N}(0,1), 
    & w_{\text{signal},i}^{(k)} = \frac{\tilde{w}_{\text{signal},i}^{(k)}}{\sum_{j=1}^{M} \tilde{w}_{\text{signal},j}^{(k)}}\\
    \tilde{w}_{\text{noise},i}^{(k)} &\sim \mathcal{N}(0,1), 
    & w_{\text{noise},i}^{(k)} = \frac{\tilde{w}_{\text{noise},i}^{(k)}}{\sum_{j=1}^{N} \tilde{w}_{\text{noise},j}^{(k)}}
\end{align}
Resulting in a linear combination of all signal components and a linear combination of all noise components for each variate.
\begin{align}
    y_{\text{signal}}^{(k)} &= y_{\text{s}}^{(k)} = \sum_{i=0}^{M} w_{\text{signal},i} \cdot y_{\text{signal},i}^{(k)} \\
    y_{\text{noise}}^{(K)} &= y_{\text{n}}^{(k)} = \sum_{i=0}^{N} w_{\text{noise},i} \cdot y_{\text{noise},i}^{(k)}
\end{align}
Each resulting time series is then z-normalized to have zero mean and unit variance.\\
In the second stage, these aggregated signal and noise series are combined into the final variate using SNR-based weights. 
Let $r^{(k)}$ denote the empirical correlation \cite{pearson1895_correlation} 
\begin{align}
    r^{(k)}_{y_{s},y_{n}} = \frac{\sum^N_{i=1}(y^{(k)}_{s} - \mu^{(k)}_{s})(y^{(k)}_{n} - \mu^{(k)}_{n})}{\sqrt{\sum^N_{i=1}(y^{(k)}_{s} - \mu^{(k)}_{s})^2}\sqrt{\sum^N_{i=1}(y^{(k)}_{n} - \mu^{(k)}_{n})^2}}
\end{align}
between $y_{\text{s}}^{(k)}$ and $y_{\text{n}}^{(k)}$, and let $\text{SNR}^{(k)}$ denote the sampled signal-to-noise ratio for variate $k$ based on a global specified snr value $\text{SNR}_{global}$ and a specified sampling variance $\sigma_{snr}^2$. 
\begin{align}
    \text{SNR}^{(k)} \sim \mathcal{N}(\text{SNR}_{global},\sigma_{snr})
\end{align}
The scaling weights are defined as
\begin{align}
    w_{\text{noise}}^{(k)} &= \frac{1}{\sqrt{1 + \text{SNR}^{(k)} + 2r^{(k)}_{y_{s},y_{n}}\sqrt{\text{SNR}^{(k)}}}} \\
    w_{\text{signal}}^{(k)} &= \sqrt{\text{SNR}^{(k)}} \cdot w_{\text{noise}}^{(k)}
\end{align}
so that the final time series satisfies both unit variance and the desired SNR: 
\begin{equation}
    y^{(k)} = w_{\text{signal}}^{(k)} \cdot y_{\text{signal}}^{(k)} 
            + w_{\text{noise}}^{(k)} \cdot y_{\text{noise}}^{(k)}
\end{equation}
This two-step procedure ensures that each variate contains a structured signal component and a controlled noise component, while allowing the difficulty of the forecasting task to be systematically adjusted through the SNR (Figure \ref{fig:component_aggregation}).

\section{Experimental Design}
We conduct a comprehensive evaluation to analyze the fundamental capabilities and limitations of four representative M-LTSF models. 
Our evaluation systematically tests these models across multiple instances of our synthetic dataset framework, where each instance is generated with specific parameter configurations to isolate and examine particular model behaviors.

\subsection{Evaluation Models}
For our evaluation, we selected four architecturally diverse M-LTSF models based on their distinct design principles and recent performance in the literature. 
These models include S-Mamba as a state-space model, iTransformer as an attention-based approach, R-Linear as a linear method, and Autoformer utilizing a decomposition-based design (see Table \ref{tab:eval_ltsf_models}).
The models differ in both size and training efficiency: Autoformer is the largest with 14.6 Million parameters but has the fastest training time of 75 seconds per epoch.
Attempts to increase Autoformer's parameters further resulted in out-of-memory errors on our smaller compute resources. 
In comparison, R-Linear has 7.5M parameters and requires 225 seconds per epoch, while S-Mamba and iTransformer have 9.2M and 6.4M parameters, with training times of 219 and 216 seconds per epoch, respectively.
The hyperparameters for each model, as well as those used in the experiments, are available in the associated GitHub repository.
\begin{table*}[htp]
\centering
\caption{Overview of evaluated M-LTSF Models\\Core Algorithm of Architecture, Seasonal Decomposition (separate modeling of trend and seasonal pattern), bi-directional Architecture with forward and inverse temporal features, Number of Parameters, Train and Inference throughput)\\}
\label{tab:eval_ltsf_models}
\begin{tabular}{@{ }l@{ }|@{ }c@{ }|@{ }c@{ }|@{ }c@{ }|@{ }c@{ }|@{ }c@{ }|@{ }c@{ }}
\toprule
\textbf{Model} & \textbf{Architectural Core} & \textbf{Seasonal Decomp.} & \textbf{Bi-directional} & \textbf{\#Parameters} & \textbf{Train [it/s]} & \textbf{Inference [it/s]}\\
\midrule
Autoformer & Correlation & \textcolor{ForestGreen}{\cmark}, running avg.  & \textcolor{Maroon}{\xmark} & 14.6M & 278.2 & 514.6  \\

iTransformer & Attention & (\textcolor{ForestGreen}{\cmark}, instance norm.) & \textcolor{ForestGreen}{\cmark} & 6.4M  & 96.6 & 325.4 \\

R-Linear & Linear & (\textcolor{ForestGreen}{\cmark}, instance norm.) & \textcolor{Maroon}{\xmark} & 7.5M & 92.7 & 1001.0 \\

S-Mamba & SSM & (\textcolor{ForestGreen}{\cmark}, instance norm.) & \textcolor{ForestGreen}{\cmark} & 9.2M  & 95.3 & 270.7 \\
\bottomrule
\end{tabular}
\end{table*}

\subsection{Evaluation Approach}
Rather than evaluating the models on a single comprehensive dataset containing all possible parameter combinations, we generate separate dataset instances, each configured to test specific aspects of model performance. 
Each dataset instance represents a unique combination of signal type, frequency band, noise type, and signal-to-noise ratio.
This controlled approach enables systematic isolation of how these factors interact and affect model performance, revealing relationships that would remain obscured in traditional benchmark evaluations on real-world datasets.\\
Each dataset instance consists of $N = 24 \cdot 365 \cdot 4 = 35040$ samples (modeling data recording of 1 sample per hour for four years) with $V = 800$ variates, providing a realistic scale for multivariate forecasting scenarios.
Across all experiments, we use a lookback horizon of $T = 96$ and the forecast horizon to $H = 96$, which reflects a realistic setup where four days of past observations are used to forecast the subsequent four days.
Each model is trained for five epochs in each individual experiment, by which point all models have shown sufficient convergence.

\subsection{Parameter Space}
Our evaluation explores the following parameter dimensions. 
For signal types, we consider three fundamental waveforms. 
The first type consists of sine waves, which provide smooth, continuously differentiable signals. 
The second type comprises smoothed sawtooth waves, characterized by linear ramps and nearly abrupt resets.
The third type includes smoothed square waves, featuring signals with sharper transitions and constant segments. 
Both square and sawtooth signals are smoothed to create jerk-free signals. 
These signal types represent fundamentally different mathematical properties and capture diverse real-world temporal patterns found in applications such as energy consumption, sensor readings, and economic indicators.\\
\begin{figure}[htp]
\begin{center}
\includegraphics[width=0.5\columnwidth]{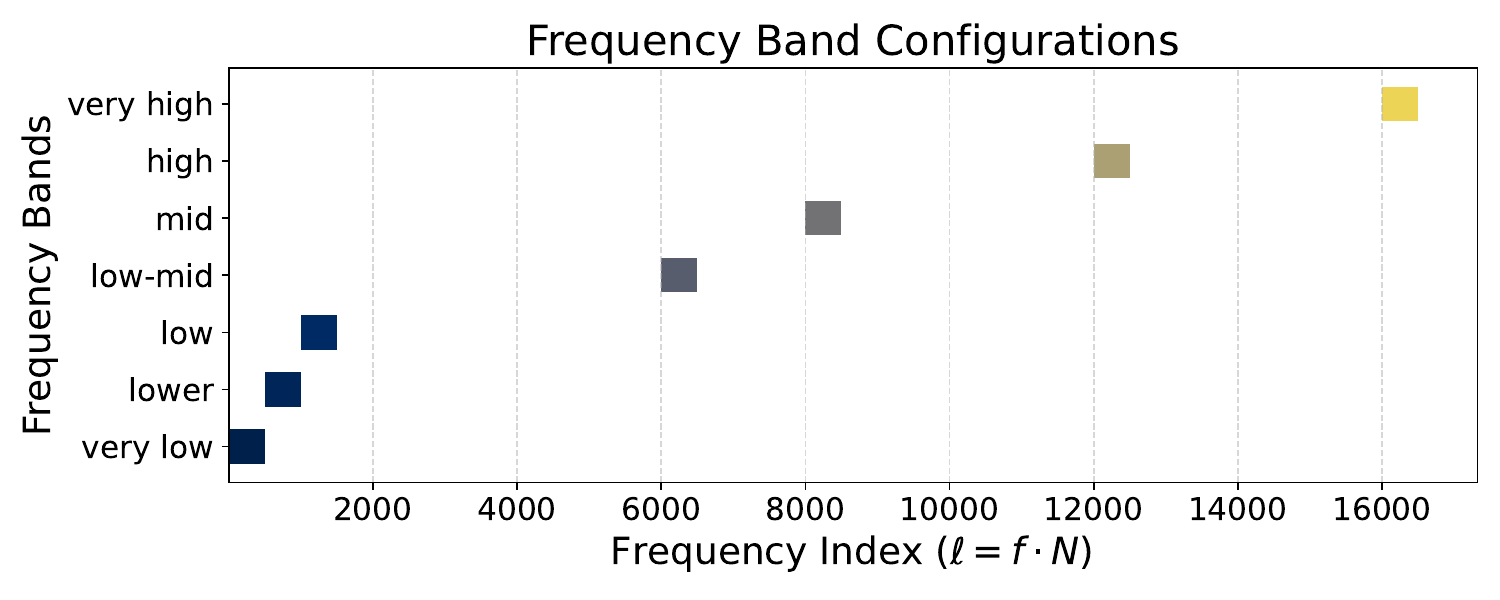}
\caption{Overview of the frequency index ranges used in the evaluations for analyzing frequency-dependent model behaviors. The defined ranges are: very low (1–500), low (500–1000), low-mid (1000–1500), mid (6000–6500), mid-high (8000-8500), high (12000–12500), and very high (16000–16500).}
\label{fig:freq_configs}
\end{center}
\end{figure}%
For frequency content, we systematically investigate model responses across seven distinct frequency index bands. The frequency index is defined as $l = f \cdot N$ where $f$ is the periodic patterns frequency and $N$ is the sequence length. 
Our frequency bands (see Figure \ref{fig:freq_configs}) include a very-low frequency index range from 1 to 500 (exhibiting near-trend behavior for short lookback horizons), a low frequency index range from 500 to 1000, a low-mid frequency index range from 1000 to 1500, a mid range from 6000 to 6500, a mid-high range from 8000-8500, a high range from 12000 to 12500, and a very high frequency index range from 16000 to 16500.
All frequency bands remain within the Nyquist sampling limit of $l_{max}=\frac{N}{2}=17520$.\\
For noise characteristics, we examine the models performance using five distinct noise types: white noise, Brownian noise, impulse noise, trend-dependent noise, and seasonal dependent noise.
\begin{figure}[htp]
\begin{center}
\includegraphics[width=0.5\columnwidth]{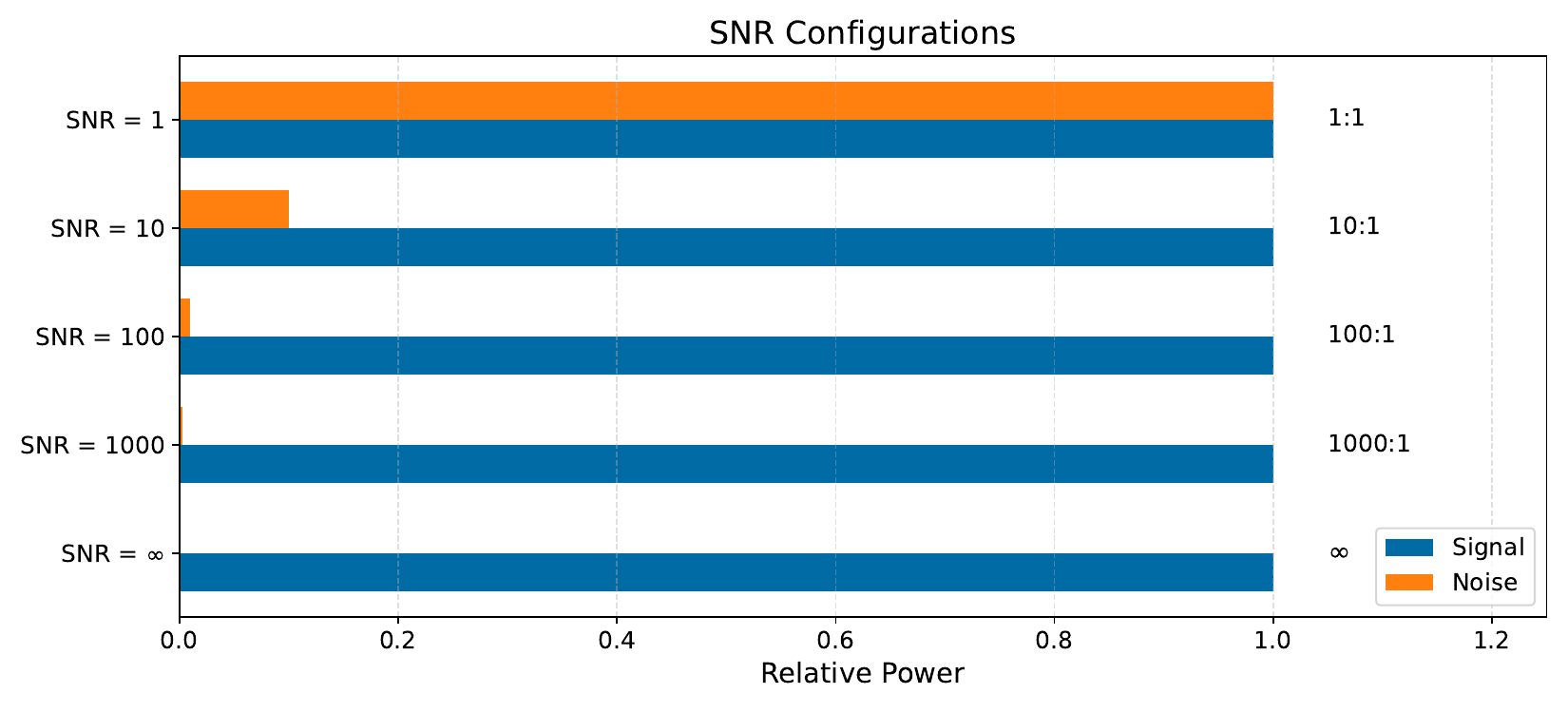}
\caption{Overview of the signal-to-noise ratios used in the evaluations for analyzing noise-dependent model behaviors. The defined ranges are: infinite SNR (no noise) as our reference, low noise with SNR of 1000 (where signal power is 1000 times greater than noise power), low-to-moderate noise with SNR of 100, moderate noise with SNR of 10, and high noise with SNR of 1 (where signal and noise have equal magnitude).}
\label{fig:snr_configs}
\end{center}
\end{figure}%
Each noise type exhibits different statistical properties and affects signal structure in fundamentally different ways.
Additionally, we vary the signal-to-noise ratio across five levels.
These include a noise-free condition with infinite SNR serving as our reference, low noise with SNR of 1000 (where signal power is 1000 times greater than noise power), low-to-moderate noise with SNR of 100, moderate noise with SNR of 10, and high noise with SNR of 1 (where signal and noise have equal magnitude).

\subsection{Dataset Instance Generation}
By combining these parameter dimensions, we generate dataset instances that capture the full factorial design space. 
For example, one instance might contain sine waves in the low frequency band with white noise at SNR of 10, while another instance combines sawtooth waves in the high frequency band with seasonal-dependent noise at SNR of 100. 
This systematic approach allows us to evaluate how models perform not just under isolated conditions, but across realistic combinations of signal characteristics, frequency content, and noise properties that better represent the complexity of real-world time series data.
Critically, for all instances containing noise, models are trained on the noised data but evaluated on the corresponding noise-free signal. This experimental design allows us to assess whether models learn the underlying signal structure or simply overfit to noise patterns, as formalized in Equation \ref{eq:test_loss}:
\begin{equation}
    \text{MSE}_{\text{test},\theta} = \frac{1}{H} \sum_{i=1}^{H} \left(y_{\text{signal},i} - f_{\theta}(x_{\text{noised},i})\right)^2 \label{eq:test_loss}
\end{equation}
This evaluation approach directly measures how well each model can recover the latent signal structure despite the presence of noise during training.

\subsection{Statistical Robustness}
To ensure robustness and account for randomness in both training and data generation, we repeat each dataset instance configuration nine times using three different random seeds for model training and three for dataset instantiation. 
This approach provides statistical confidence in our findings and ensures that observed performance differences reflect genuine model characteristics rather than random variations.\\
By systematically generating and evaluating models on these parameterized dataset instances (each representing a specific combination of signal type, frequency band, noise type, and noise level), our evaluation framework provides a consistent and informative basis for comparing M-LTSF models. 
This approach enables in-depth understanding of which specific conditions and parameter combinations may cause systematic biases or failures in different model architectures.

\section{Evaluation}
This chapter presents a comprehensive evaluation of the four selected M-LTSF models across our systematically generated synthetic dataset instances. 
By controlling and isolating specific signal characteristics, we can identify fundamental capabilities and limitations of each model architecture that are difficult to disentangle in real-world benchmark datasets.

\subsection{Noise and Trend-Free Baseline Evaluation}
\begin{figure}[htp]
\begin{center}
\includegraphics[width=0.7\columnwidth]{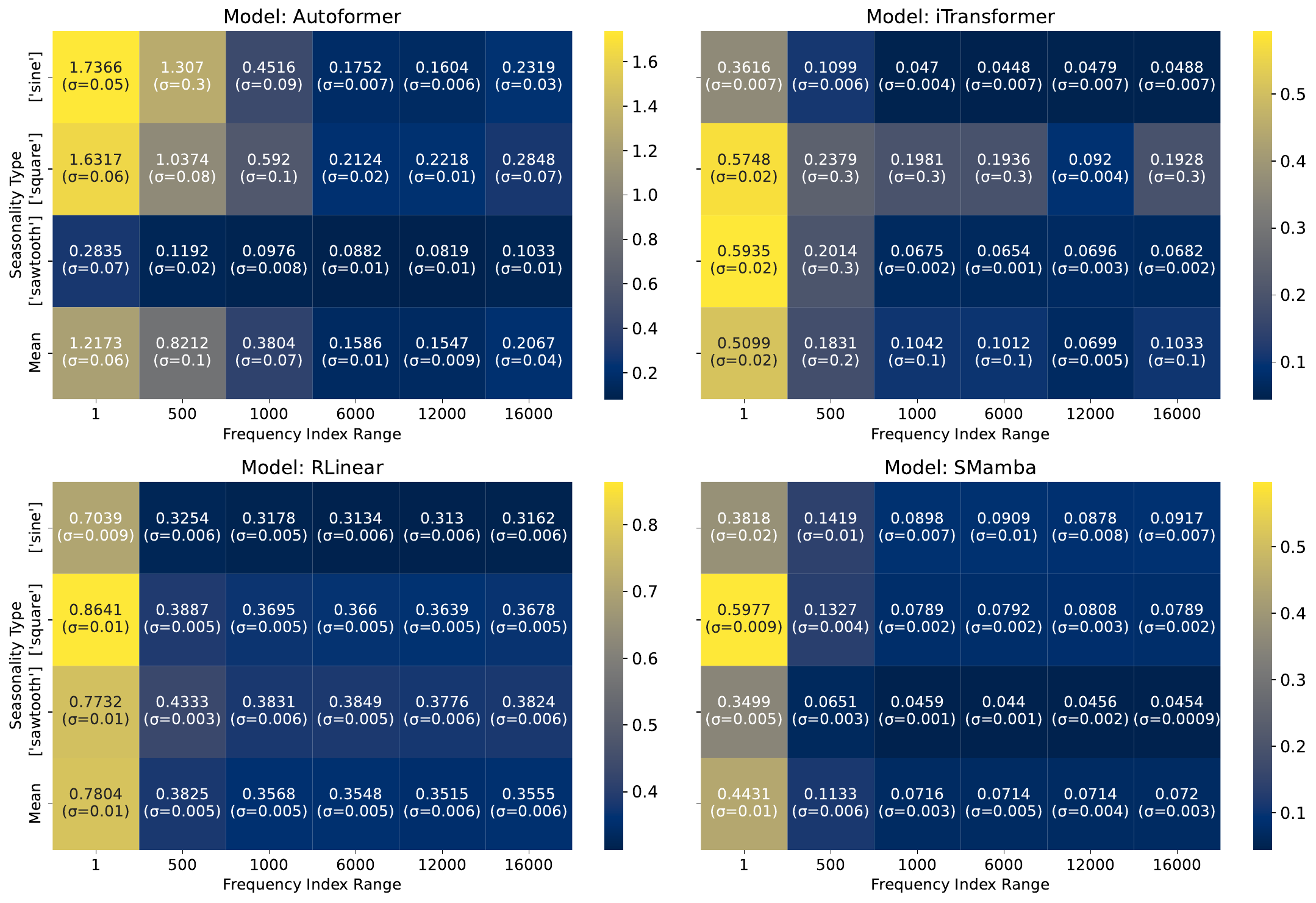}
\caption{MSE heatmap across the four different M-LTSF models (Autoformer. R-Linear, S-Mamba, iTransformer) across noise-free and trend-free signals across multiple frequency ranges (1–500, 500–1000, 1000–1500, 6000–6500, 12000–12500, 16000–16500) and signal types (sine, square, sawtooth). Autoformer and S-Mamba show their respective best performance on sawtooth-like waves while iTransformer and R-Linear show better performance on sinusoidal data. Please not the different scales for each individual model. Blue indicates lower MSE (better performance), while yellow indicates higher MSE (worse performance).}
\label{fig:no_noise_no_trend_heatmaps}
\end{center}
\end{figure}%
Model performance was first evaluated under ideal conditions without noise or trend components across multiple frequency ranges (1-500, 500-1000, 1000-1500, 6000-6500, 12000-12500, 16000-16500) and signal types (sine, square, sawtooth).\\
The results shown in Figure \ref{fig:no_noise_no_trend_heatmaps} reveal a critical limitation when the frequency of periodic patterns becomes too small relative to the lookback window. 
When the lookback window is smaller than the shortest period in the signal, model performance decreases significantly across all architectures. 
In average the MSE decreased about 0.575 between the best performing frequency range (6000-6500) and the lowest frequency range (1-500) over all models and seasonal types.
This occurs since the models cannot capture complete periodic cycles within their input horizon, causing them to misinterpret periodic behavior as trend components.
The lowest frequency index that can be captured by the lookback window of $T=96$ is $l = \lceil \frac{1}{T} \cdot N \rceil = 355$. 
Consequently, some of the frequencies in the lowest range of 1 to 500 fall below this threshold resulting in strong degradation of performance.\\
Interestingly, no significant performance differences are observed between higher frequency ranges, suggesting that once the frequencies are high enough for the look back window of $T=96$ the model can adequately captures periodic patterns.
Autoformer shows some performance differences in the lower-mid frequency ranges, since the seasonal-decomposition is highly dependent on the window size used for calculation of the running mean.
Thus lower frequencies have more impact than higher frequency ranges.\\
In addition, the models demonstrate distinct preferences for different signal types. 
Autoformer and S-Mamba achieve their respective best performance on sawtooth-like seasonal patterns.
As seen in Figure \ref{fig:no_noise_no_trend_heatmaps}, the Autoformer architecture improves on MSE from 0.1752 and 0.2124, for sine and square seasonal types respectively, to a MSE of 0.0882 for a frequency range of 6000-6500. 
Possibly due to its trend-seasonal decomposition mechanism that can isolate the sawtooth waves into its seasonal component and short trend component for each period.
For S-Mamba the performance improves from a MSE of 0.0869 and 0.0792 for the other signal types to a MSE of 0.044 when signals are built using smoothed sawtooth waves.
This could be explained by S-Mambas time-varying state-space representation where the gating mechanism on the input and the time-varying state transition excels on modeling the sharp transitions of a sawtooth wave.\\
R-Linear and iTransformer achieve their respective best performance on sine-like seasonal patterns.
R-Linears performance only increases in MSE by a slight relative margin from 0.366 and 0.3849, for square and sawtooth seasonal types respectively, to a MSE of 0.3134 at a frequency range of 6000-6500.
Since the dataset has a high number of variates with complex cross-dependencies, R-Linear does not have enough expressivity to model these complex dynamics by channel-independent linear transformations.
This results in the general poor performance of R-Linear in this setting.
One possible explanation for iTransformers better performance on sine wave data is that the self-attention mechanism may be better at capturing the gradual phase relationships inherent in smooth sinusoidal patterns, whereas the sharper transitions in sawtooth and square waves may interfere with the attention weights ability to establish coherent temporal relationships across the sequence.\\

\subsection{Trend Component Impact}
\begin{figure}[htp]
\begin{center}
\includegraphics[width=0.6\columnwidth]{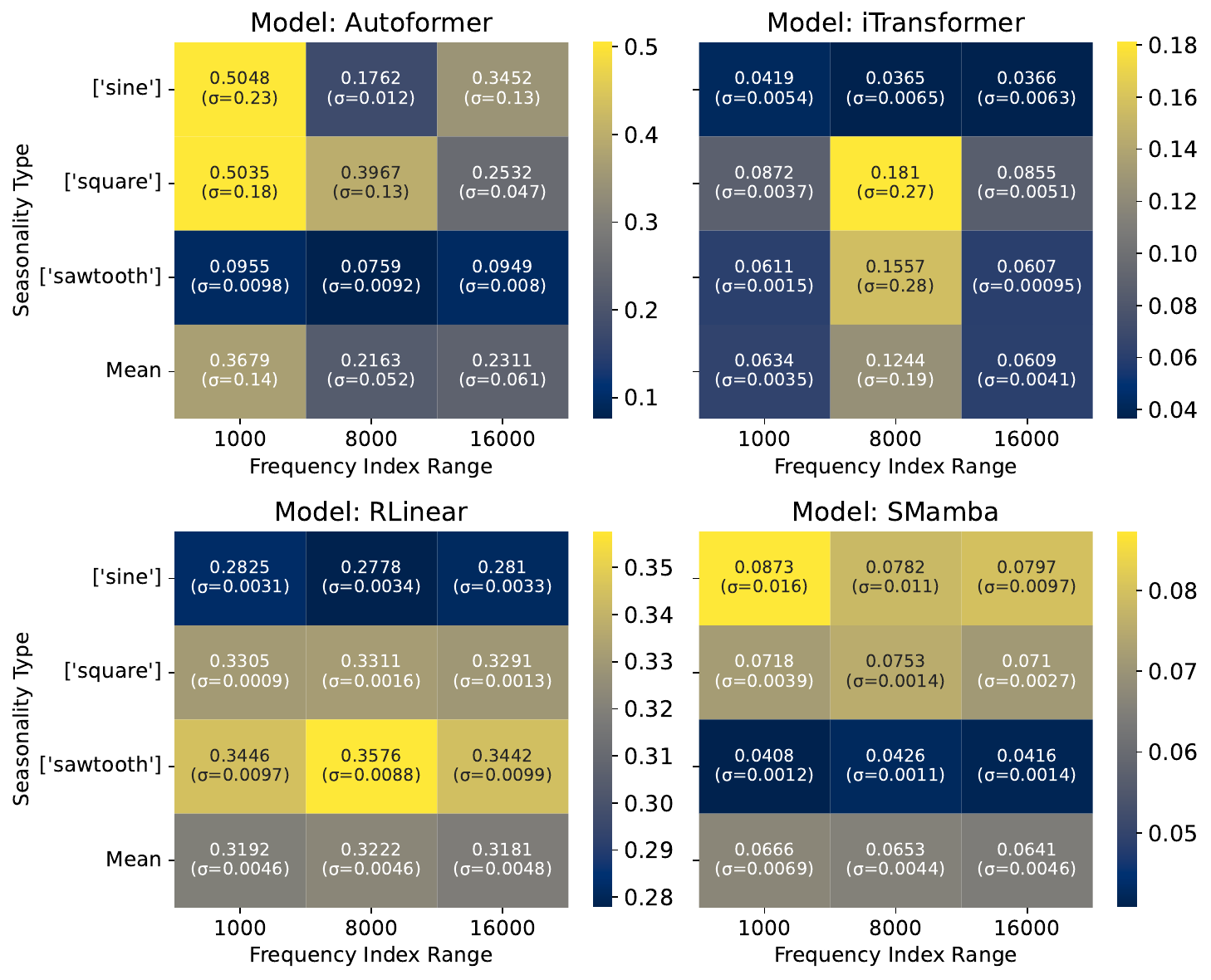}
\caption{MSE heatmap for noise-free signal with trend components across multiple frequency ranges (1000-1500, 8000-8500, 16000-16500) and signal types (sine, square, sawtooth). Compared to the no-trend case (Figure \ref{fig:no_noise_no_trend_heatmaps}), all models demonstrate improved performance while maintaining consistent seasonality type preferences. Please not the different scales for each individual model. Blue indicates lower MSE (better performance), while yellow indicates higher MSE (worse performance).}
\label{fig:no_noise_with_trend_heatmaps}
\end{center}
\end{figure}%

To get a better understanding how trend components in the signal influence model performances, trend components were introduced to the synthetic data for a smaller subset of frequencies (frequency ranges: 1000-1500, 8000-8500, 16000-16500) to balance meticulousness and computational effort.
A consistent improvement in model performance across all architectures compared to purely seasonal signals (see Figure \ref{fig:no_noise_with_trend_heatmaps}) could be observed.
This improvement can possibly be attributed to the directional guidance that trend components provide during gradient-based optimization. 
Purely seasonal signals can create challenging optimization landscapes due to their inherent symmetries and the phase prediction requirements that arise when models must learn to match periodic patterns precisely. 
The trend component serves as an "anchor" that helps models establish temporal direction and reduces the phase prediction burden that is particularly challenging in purely periodic signals.
The relative performance rankings between models remain consistent with trend addition, indicating that the fundamental signal type preferences are architectural characteristics rather than artifacts of the no-trend condition.

\subsection{Noise Robustness Analysis}
The noise robustness of the evaluated architectures was examined under five noise types (white, Brownian, impulse, trend and seasonal noise) at multiple signal-to-noise ratios ($\infty$ 1000, 100, 10 and 1) and across two frequency index ranges (1000-1500 and 16000-16500).
The evaluation reveals systematic differences on how models respond to noise as well as distinct architectural weaknesses.\\
\begin{figure}[htp]
\begin{center}
\includegraphics[width=0.6\columnwidth]{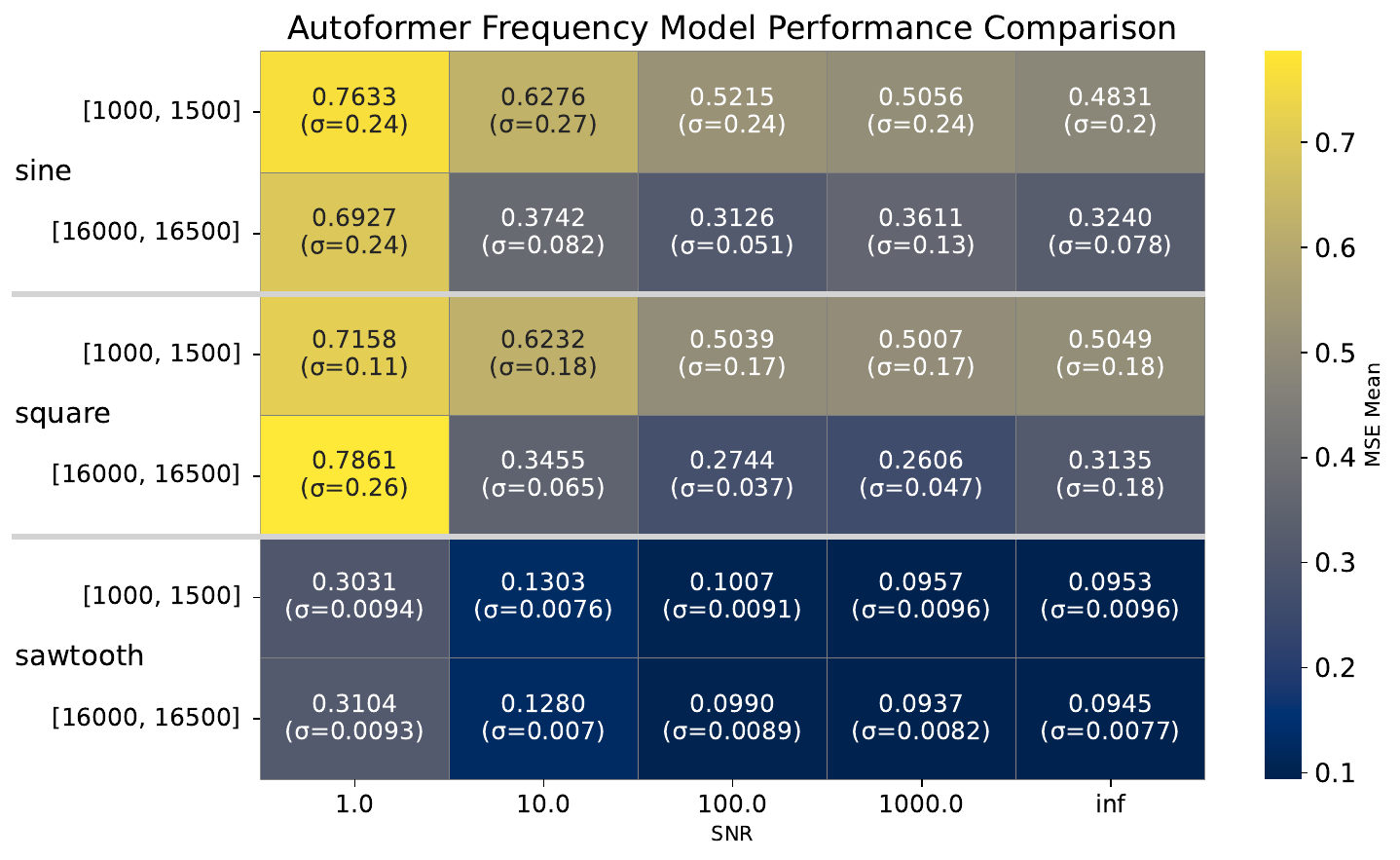}
\caption{MSE heatmap of Autoformer performance across two frequency index ranges (1000-1500, 16000-16500) and multiple signal-to-noise ratios ($\infty$, 1000, 100, 10, 1). Signals are degraded with additive white noise. For sine and square-wave data, Autoformers performance improves significantly with higher frequencies (MSE difference of 0.2089 and 0.2295 for sine and square respectively at a SNR value of 100). For sawtooth-like seasonality types the model only improves by a small margin. Blue indicates lower MSE (better performance), while yellow indicates higher MSE (worse performance).}
\label{fig:heatmap_autoformer}
\end{center}
\end{figure}%
As seen in Figure \ref{fig:heatmap_autoformer}, performance across the two evaluated frequency index ranges showed that for the higher frequencies Autoformers MSE improved when introduced to noise and seasonal types of sine and square, while for the sawtooth type its error remained similar between the two frequency ranges.
The other architectures exhibited largely consistent performance across frequency ranges, suggesting a less frequency-selective behavior.\\
The analysis of noise type and signal-to-noise ratio (SNR) on model performance showed different model behavior.
\begin{figure}[htp]
\begin{center}
\includegraphics[width=0.6\columnwidth]{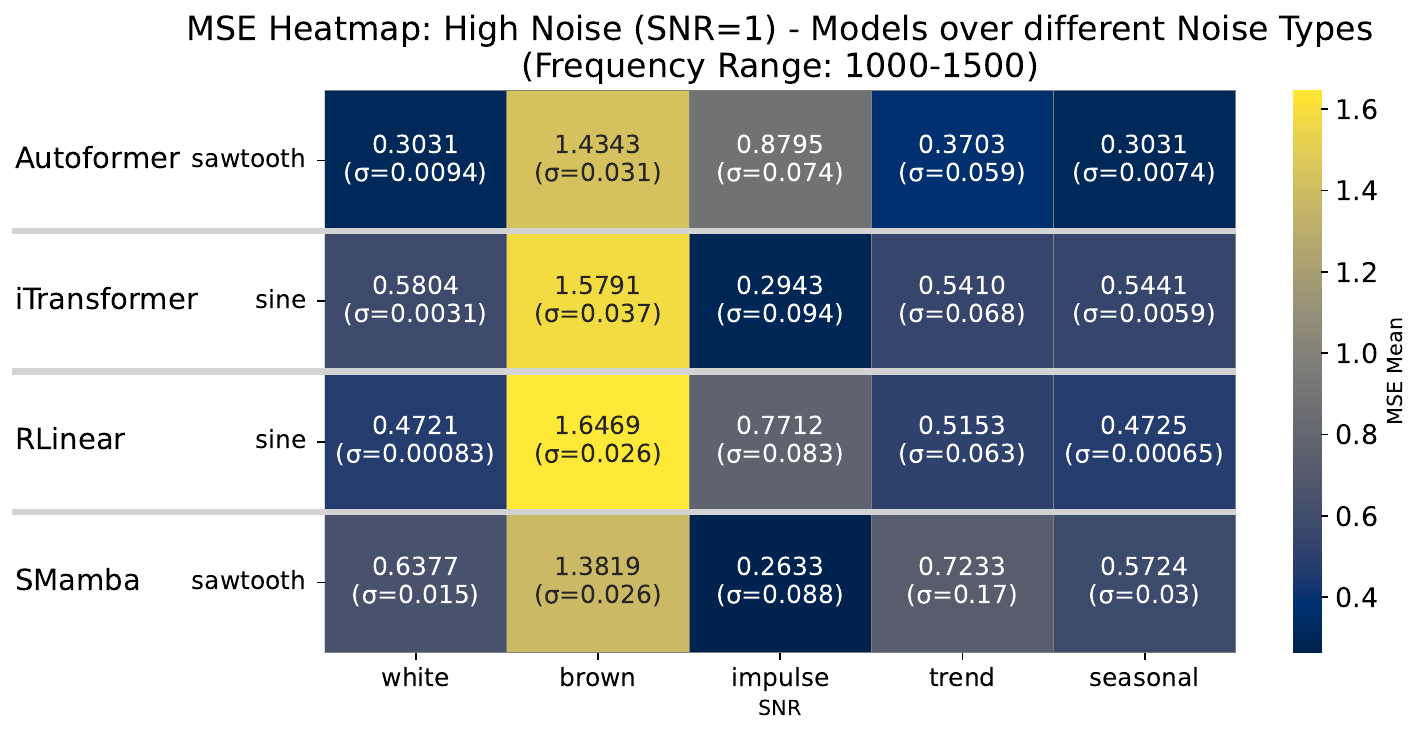}
\caption{MSE heatmap of model performance under the highest noise setting (SNR=1) across different noise types (white, Brownian, impulse, trend, seasonal). The evaluation considers the frequency range 1000–1500, with each individual architectures best seasonality type selected. Across all noise types, Brownian noise with a SNR value of 1 shows the largest degradation in MSE across all architectures. Blue indicates lower MSE (better performance), while yellow indicates higher MSE (worse performance).}
\label{fig:heatmap_high_noise}
\end{center}
\end{figure}%
Brownian noise caused the strongest degradation at the lowest SNR of 1 (Figure \ref{fig:heatmap_high_noise}), across all models, highlighting the challenge of modeling non-stationary noise. 
For SNR values between 10 and 100, performance improved for all noise types drastically, reflecting the presence of sufficient clean data for models to capture underlying dynamics.
\begin{figure}[htp]
\begin{center}
\includegraphics[width=0.6\columnwidth]{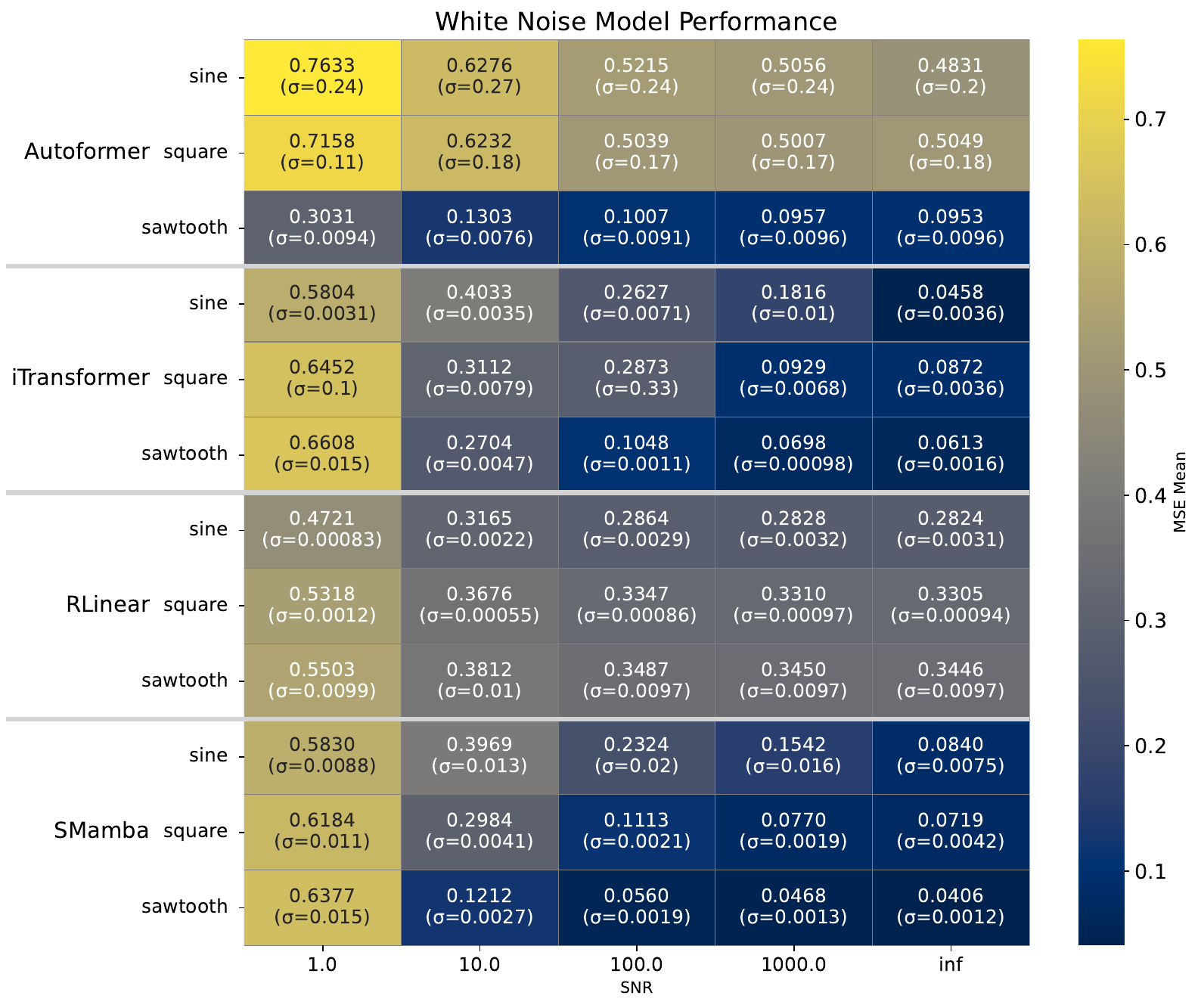}
\caption{MSE heatmap of model performance under white noise across all seasonal types (sine, square, sawtooth) and varying SNR values ($\infty$, 1000, 100, 10, 1). The evaluation considers the frequency range of  1000-1500. Across all architectures the MSE increases the lower the SNR value (e.g. S-Mamba has a difference in MSE of 0.5971 for sawtooth seasonal patterns between SNR values of infinity and 1). The model-specific best seasonality types are persistent even with noise. Blue indicates lower MSE (better performance), while yellow indicates higher MSE (worse performance).}
\label{fig:heatmap_white_noise}
\end{center}
\end{figure}%
White noise (shown in Figure \ref{fig:heatmap_white_noise}) produced the expected trend in model performance for all architectures, with MSE decreasing as SNR increased, confirming that cleaner signals facilitate learning of the true system dynamics.
Notably, the signal type preferences observed in noise-free conditions persist even under various noise conditions, indicating that these preferences are fundamental architectural characteristics rather than artifacts of clean data conditions.\\
Architectural characteristics further shaped sensitivity to noise. 
\begin{figure}[htp]
\begin{center}
\includegraphics[width=0.6\columnwidth]{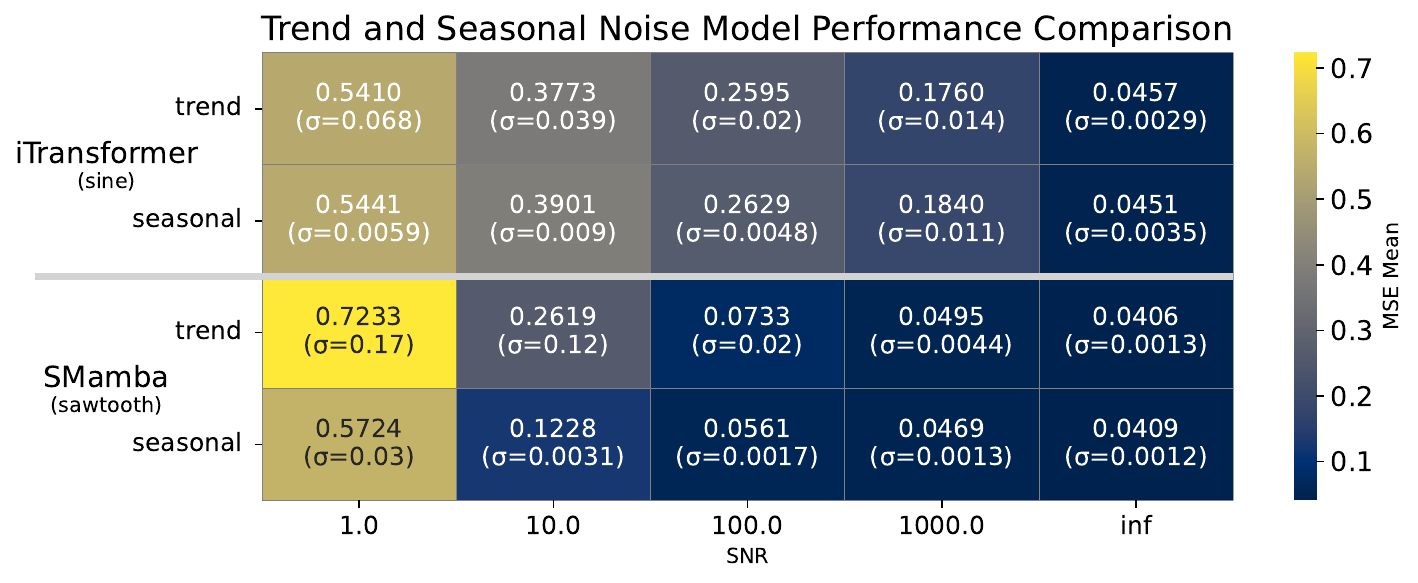}
\caption{MSE heatmap comparing trend and seasonal noise between S-Mamba and iTransformer across different SNR values ($\infty$, 1000, 100, 10, 1). The evaluation considers the frequency range 1000–1500, with each individual architectures best seasonality type selected. S-Mamba shows a higher sensitivity to trend noise in respect to seasonal noise, while iTransformer shows a higher sensitivity to seasonal noise. Blue indicates lower MSE (better performance), while yellow indicates higher MSE (worse performance).}
\label{fig:heatmap_trend_seasonal_noise}
\end{center}
\end{figure}%
Additionally as shown in Figure \ref{fig:heatmap_trend_seasonal_noise}, iTransformer was more vulnerable to seasonal noise, whereas S-Mamba exhibited higher sensitivity to trend noise, reflecting the differences of attention-based and state-space designs. 
R-Linear, on the other hand, reached performance saturation even at high SNR values, revealing limited expressivity in capturing complex dependencies across many variates.
Autoformer exhibits high performance variability on sine and square signals, likely due to hyperparameter sensitivity in its decomposition mechanism, which relies on the running average window size.
However, it shows robustness against white, trend, and seasonal noise when processing sawtooth signals, suggesting that its trend-seasonal decomposition aligns well with the sharp transitions in sawtooth patterns.\\
These observations emphasize that noise impacts are not uniform and are strongly dependent on the interplay between signal type, frequency content, noise characteristics, and model architecture.
The evaluation demonstrates that the ability of M-LTSF models to maintain performance under noisy conditions is highly architecture-dependent, with specific strengths and vulnerabilities emerging under different types of noise and frequency scenarios, highlighting the importance of considering both signal and noise properties when selecting or designing forecasting models.

\subsection{Spectral Analysis}
To complement the noise robustness evaluation, we conducted a spectral analysis to examine how well M-LTSF models preserve frequency domain characteristics of the original signal. 
The MSE is a commonly used metric for evaluating model predictions.
However, averaging it over all time steps can obscure how prediction errors develop across the prediction horizon, as errors are typically smaller at the beginning of the horizon and grow over time~\cite{zhou2025_dynamicalerrors}.
Spectral analysis offers a complementary perspective by assessing not only the difference between prediction and ground truth but also whether the model has accurately captured the true spectral components of the signal.
This approach provides deeper insight into frequency-specific behaviors and highlights inherent limitations in current architectures in learning and maintaining accurate spectral representations over extended prediction horizons.\\
The spectral analysis reveals that none of the evaluated architectures successfully learns a clean spectrum representation, even under ideal noise-free conditions (SNR = $\infty$).
This finding aligns with the expected behavior of gradient descent optimization in deep learning, where the algorithm typically converges to local minima rather than the global minimum which in our setting introduces spurious frequency components.\\
\begin{figure}[htp]
\begin{center}
\includegraphics[width=0.6\columnwidth]{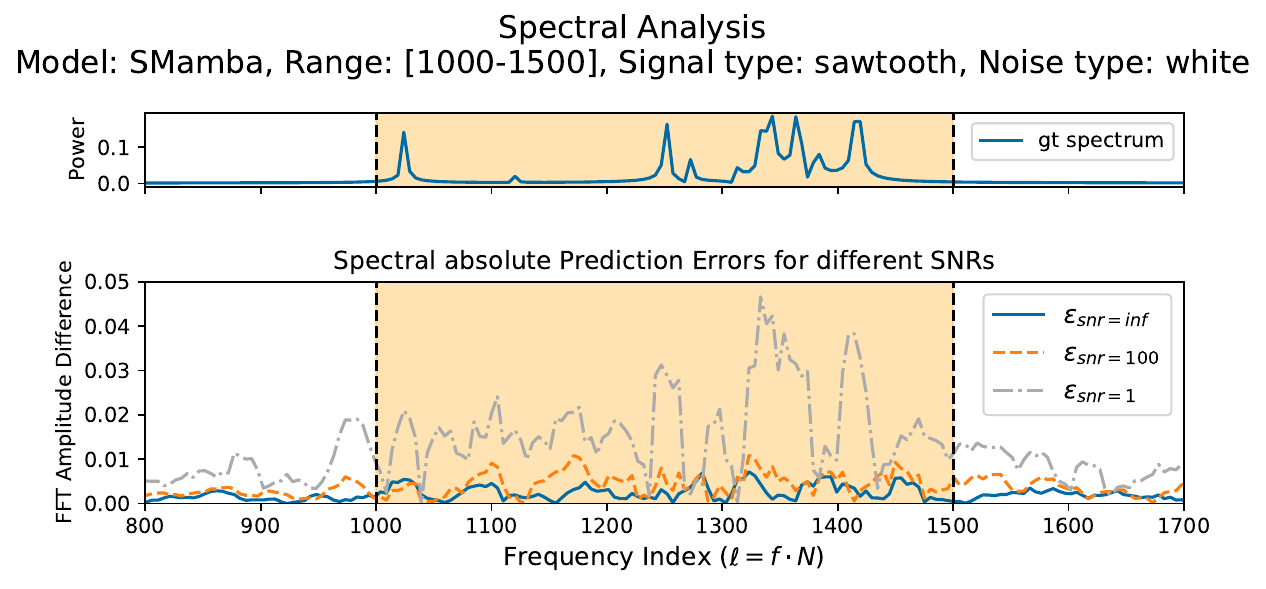}
\caption{Spectral Analysis of a S-Mamba model trained on synthetic data based on sawtooth waves and a frequency index range of 1000-1500 with added white noise. The upper graph shows the spectrum of the data of one variate without noise. The lower graph shows the absolute prediction error $\epsilon$ between the ground truth spectrum as seen in the upper graph and the predicted spectrum based on different SNR values ($\infty$, 100, 1).}
\label{fig:spectral_smamba}
\end{center}
\end{figure}%
S-Mamba performance shown in Figure \ref{fig:spectral_smamba} demonstrates the most promising spectral learning capabilities among the evaluated models. 
At infinite SNR, S-Mamba produces minimal error between predicted and ground truth spectra, with the lowest overall spectral error across the frequency range of 1000-1500. 
As noise increases (SNR = 100), S-Mamba shows moderate spectral degradation while maintaining relatively low error levels. 
However, at high noise levels (SNR = 1), significant deviations emerge, particularly at frequencies with high power of the ground truth spectrum. 
This progressive spectral degradation directly correlates with the MSE performance patterns observed in the noise robustness evaluation, providing a frequency-domain explanation for S-Mambas superior performance under clean conditions and gradual deterioration with increasing noise.\\
\begin{figure}[htp]
\begin{center}
\includegraphics[width=0.6\columnwidth]{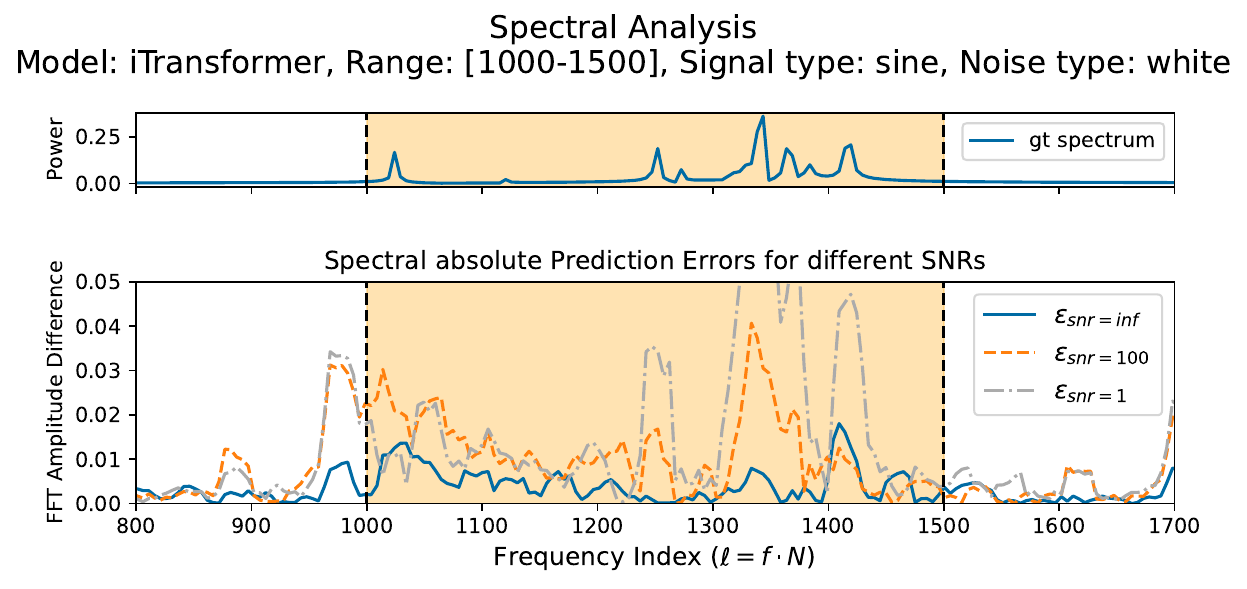}
\caption{Spectral Analysis of an iTransformer model trained on synthetic data based on sawtooth waves and a frequency index range of 1000-1500 with added white noise. The upper graph shows the spectrum of the data of one variate without noise. The lower graph shows the absolute prediction error $\epsilon$ between the ground truth spectrum as seen in the upper graph and the predicted spectrum based on different SNR values ($\infty$, 100, 1).}
\label{fig:spectral_itransformer}
\end{center}
\end{figure}%
iTransformer's spectral analysis (Figure \ref{fig:spectral_itransformer}) reveals frequency-dependent sensitivity that aligns with its noise robustness characteristics. 
The model shows clear degradation at main spectral frequencies as SNR decreases, with lower SNR values producing larger differences between predicted and ground truth spectra. 
Interestingly, at lower frequencies, iTransformer performs worse at a SNR of 100 than SNR of 1, suggesting complex interactions between the attention mechanism and frequency content under different noise conditions. 
The overall higher spectral error compared to S-Mamba corresponds well with the generally higher MSE values observed in the previous chapter (seen in Figure \ref{fig:heatmap_white_noise}).\\
\begin{figure}[htp]
\begin{center}
\includegraphics[width=0.6\columnwidth]{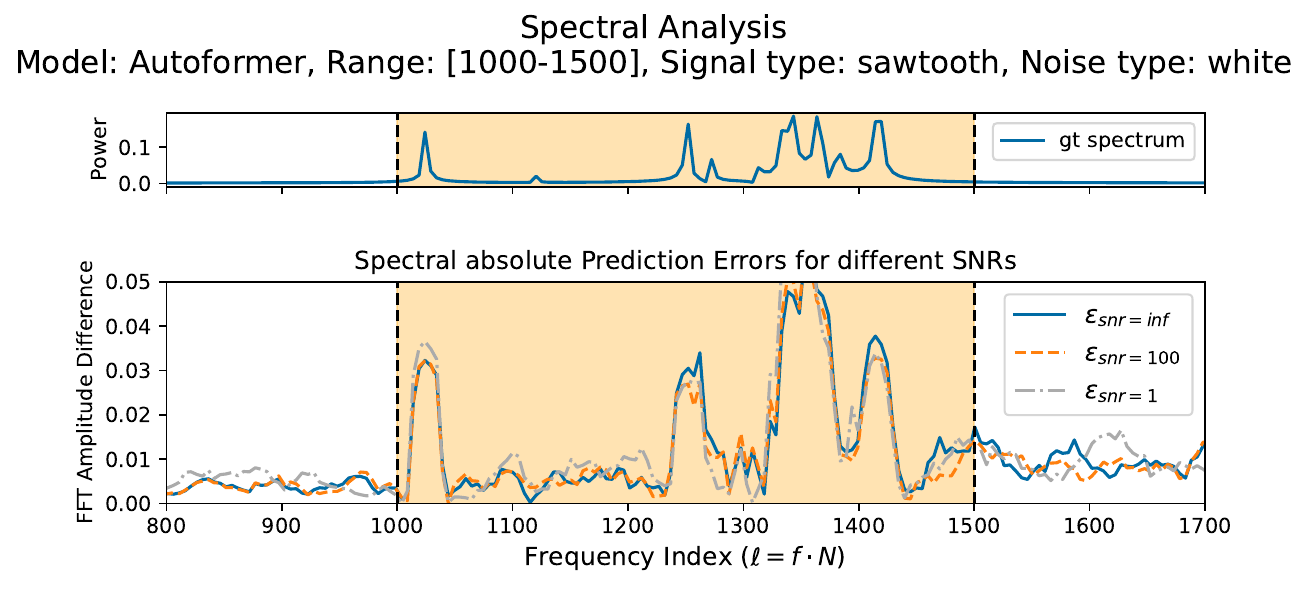}
\caption{Spectral Analysis of an Autoformer model trained on synthetic data based on sawtooth waves and a frequency index range of 1000-1500 with added white noise. The upper graph shows the spectrum of the data of one variate without noise. The lower graph shows the absolute prediction error $\epsilon$ between the ground truth spectrum as seen in the upper graph and the predicted spectrum based on different SNR values ($\infty$, 100, 1).}
\label{fig:spectral_autoformer}
\end{center}
\end{figure}%
Both Autoformer and R-Linear exhibit surprisingly similar spectral predictions across different SNR values, indicating limited sensitivity to noise levels in the frequency domain. 
Autoformer shows its highest error at the main spectral frequencies with significant power, yet this poor spectral representation does not consistently translate to proportionally poor MSE performance, possibly due to the high variance observed in Autoformers MSE results. 
This disconnect between spectral accuracy and time-domain performance suggests that Autoformers decomposition mechanism may compensate for spectral inaccuracies through its trend-seasonal separation approach. 
\begin{figure}[htp]
\begin{center}
\includegraphics[width=0.6\columnwidth]{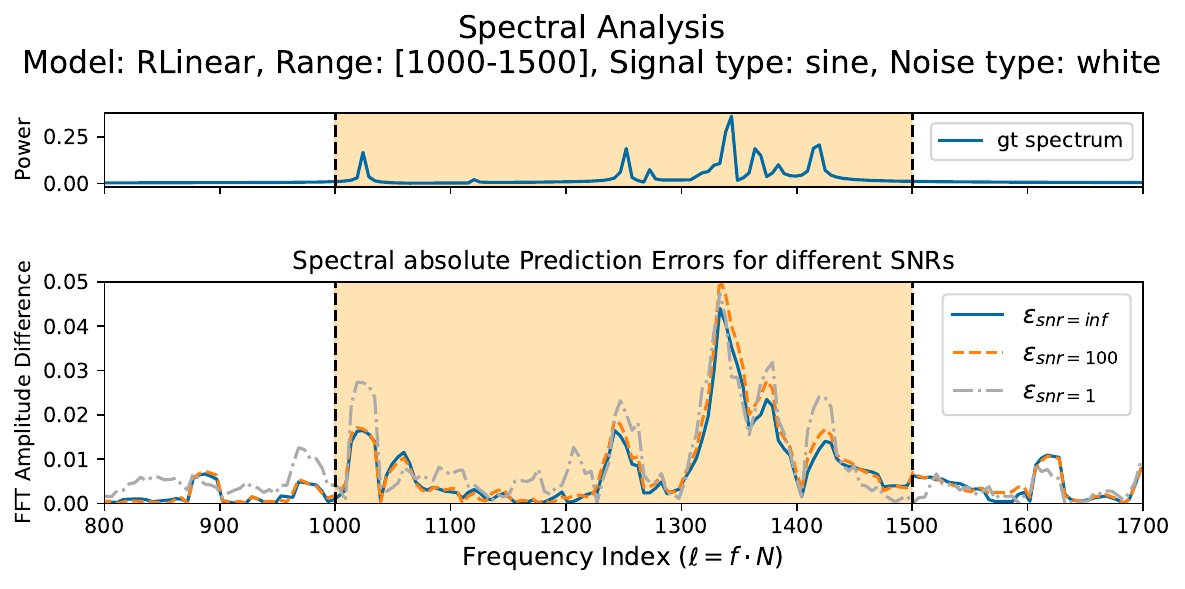}
\caption{Spectral Analysis of an R-Linear model trained on synthetic data based on sawtooth waves and a frequency index range of 1000-1500 with added white noise. The upper graph shows the spectrum of the data of one variate without noise. The lower graph shows the absolute prediction error $\epsilon$ between the ground truth spectrum as seen in the upper graph and the predicted spectrum based on different SNR values ($\infty$, 100, 1).}
\label{fig:spectral_rlinear}
\end{center}
\end{figure}%
R-Linear's consistent spectral errors across SNR values align with its observed performance saturation in the noise robustness analysis, indicating that the model operates at or near its representational capacity even under low noise conditions.\\
These spectral analysis findings provide crucial insights into the frequency-domain mechanisms underlying the noise robustness patterns observed previously. The correlation between spectral learning quality and MSE performance, particularly evident in S-Mamba and iTransformer, suggests that improving spectral representation capabilities could be a promising direction for enhancing M-LTSF model performance. Furthermore, the universal failure to achieve clean spectral representations highlights a fundamental optimization challenge in current deep learning approaches to time series forecasting that warrants further investigation.

\subsection{Model Selection}
Based on the previous experiments, the choice of M-LTSF model should be guided by specific use case requirements and data characteristics. 
An Overview to help guide selection is shown in Figure \ref{fig:model_selection}. 
R-Linear, despite not achieving the best performance in our multivariate synthetic dataset with 800 variates, serves as an effective baseline model due to its simplicity. 
Its channel-independent approach, while limiting its ability to capture cross-dependencies between time series, makes it suitable for scenarios requiring fast inference or when the dataset only contains a small number of variates.\\
Autoformer demonstrates strong performance in high-frequency scenarios and where data builds upon sawtooth-like patterns (Figure \ref{fig:heatmap_autoformer}) and exhibits effective white noise filtering capabilities through its decomposition mechanism, making it well-suited for applications with high-frequency seasonal patterns. 
However, its higher sensitivity to hyperparameters and substantial performance variability across experimental runs necessitate careful tuning and may limit its applicability in production environments where consistent performance is critical.\\
S-Mamba and iTransformer emerge as the most robust choices for complex forecasting tasks, demonstrating superior spectral learning capabilities and accurate frequency spectrum reconstruction under clean conditions (Figures \ref{fig:heatmap_white_noise},\ref{fig:spectral_smamba},\ref{fig:spectral_itransformer}). 
S-Mamba shows particular vulnerability to trend-dependent noise, suggesting it may be less suitable for datasets with systematic long-term deviations, while iTransformer maintains more consistent performance across different noise types (Figure \ref{fig:heatmap_trend_seasonal_noise}).
For applications requiring reliable spectral reconstruction and handling of diverse noise conditions, iTransformer and S-Mamba represent the most suitable choices, with the specific selection depending on the predominant noise characteristics expected in the target domain.

\section{Conclusion}
This work presented a systematic evaluation of multivariate long-term time series forecasting (M-LTSF) models using a novel parameterizable synthetic dataset framework designed to isolate and analyze specific architectural behaviors under controlled conditions. 
Our synthetic dataset approach enabled precise manipulation of signal characteristics, noise types, and signal-to-noise ratios, providing insights that remain obscured in traditional benchmark evaluations on real-world datasets.\\
The experimental evaluation of four representative M-LTSF architectures (S-Mamba, iTransformer, R-Linear, and Autoformer) revealed several fundamental limitations and architectural characteristics. 
Most critically, we demonstrated that when the lookback window is insufficient to capture complete periodic cycles, all models suffer significant performance degradation, with average MSE increases of approximately 0.575 between optimal and suboptimal frequency ranges. 
S-Mamba and Autoformer perform best on sawtooth patterns, while R-Linear and iTransformer favor sinusoidal signals.\\
Our noise robustness analysis exposed distinct architectural vulnerabilities: iTransformer showed particular sensitivity to impulse noise and seasonal dependencies, S-Mamba exhibited higher vulnerability to trend-dependent noise, while Autoformer demonstrated high performance variability despite its decomposition-based design. 
Importantly, signal type preferences observed under clean conditions persisted across various noise scenarios, indicating that these preferences represent fundamental architectural characteristics rather than artifacts of specific experimental conditions.\\
The spectral analysis revealed a universal failure across all evaluated architectures to achieve clean spectral representations, even under ideal noise-free conditions.
This finding suggests fundamental optimization challenges in current deep learning approaches to time series forecasting, where gradient-based methods converge to local minima that introduce spurious frequency components. 
Additionally, S-Mamba and iTransformer demonstrated superior frequency reconstruction capabilities under clean conditions.\\
These findings have important implications for both model selection and future research directions.\\
For practitioners, the choice of M-LTSF model should be guided by specific use case requirements: R-Linear serves as an effective baseline for simple scenarios, Autoformer excels in high-frequency sawtooth-like patterns, while S-Mamba and iTransformer represent the most robust choices for complex multivariate forecasting tasks with different noise sensitivities.\\
For researchers, our work highlights several promising directions. 
The synthetic framework could be extended with additional noise characteristics, such as colored noise with specific spectral properties, while other emerging architectures could be evaluated within this controlled setting. 
Furthermore, bridging the gap between controlled evaluation and practical applications through validation on real-world datasets with well-characterized noise properties would strengthen the generalizability of these findings. 
Our controlled evaluation framework successfully revealed behavioral differences that traditional benchmarks fail to capture, establishing a foundation for more informed model development in multivariate long-term time series forecasting.
To facilitate reproducibility and further research, the complete implementation of our synthetic dataset framework, along with all experimental configurations and hyperparameters used in this study, is publicly available.

\bibliographystyle{splncs04}
\bibliography{references}  

\end{document}